\documentclass[lettersize,journal]{IEEEtran}
\usepackage{algorithmic}
\usepackage{array}
\usepackage{url}
\usepackage{verbatim}
\usepackage{cite}
\usepackage{amsmath,amssymb,amsfonts}
\usepackage{makecell}

\usepackage{algorithm}
\usepackage{graphicx}
\usepackage{textcomp}
\usepackage{xcolor}
\usepackage{adjustbox}
\usepackage{multirow}
\usepackage{colortbl}
\usepackage{subcaption}

\usepackage{caption}
\usepackage{enumerate}
\usepackage{booktabs} 
\usepackage{pifont}

\usepackage{quoting}

\usepackage{amsmath,amsfonts,bm}

\def\eqref#1{Eq.~(\ref{#1})}

\def\1{\bm{1}}

\def\vx{{\bm{x}}}

\DeclareMathAlphabet{\mathsfit}{\encodingdefault}{\sfdefault}{m}{sl}
\SetMathAlphabet{\mathsfit}{bold}{\encodingdefault}{\sfdefault}{bx}{n}

\def\gD{{\mathcal{D}}}

\def\gT{{\mathcal{T}}}

\def\gX{{\mathcal{X}}}
\def\gY{{\mathcal{Y}}}

\newcommand{\E}{\mathbb{E}}
\newcommand{\Ls}{\mathcal{L}}

\DeclareMathOperator{\tr}{tr}

\begin{document}

\title{Shortcuts Everywhere and Nowhere: Exploring Multi-Trigger Backdoor Attacks}

\author{Yige Li, Jiabo He, Hanxun Huang, Jun Sun, Xingjun Ma, and Yu-Gang Jiang,~\IEEEmembership{Fellow, ~IEEE}
\thanks{Yige Li, Xingjun Ma, and Yu-Gang Jiang are with the Shanghai Key Lab of Intell. Info. Processing, School of CS, Fudan University, Shanghai, China (e-mail: xdliyige@gmail.com, xingjunma@fudan.edu.cn, ygj@fudan.edu.cn). }%
    \thanks{Jiabo He is with Bosch Research Asia Pacific \& Bosch Center for Artificial Intelligence (BCAI), Shanghai, China (e-mail: jiabohe1208@gmail.com).}
    \thanks{Hanxun Huang is with the School of Computing and Information Systems, the University of Melbourne, Australia (e-mail: hanxun@unimelb.edu.au).}
    \thanks{Jun Sun is with the School of Computing and Information Systems, Singapore Management University, Singapore (e-mail:junsun@smu.edu.sg).}
    \thanks{The work was partially completed during Yige's internship at Bosch Research Asia Pacific \& BCAI. Corresponding Author: Xingjun Ma.}
}



\maketitle

\begin{abstract}
Backdoor attacks have become a significant threat to the pre-training and deployment of deep neural networks (DNNs). Although numerous methods for detecting and mitigating backdoor attacks have been proposed, most rely on identifying and eliminating the ``shortcut" created by the backdoor, which links a specific source class to a target class. However, these approaches can be easily circumvented by designing multiple backdoor triggers that create shortcuts everywhere and therefore nowhere specific. In this study, we explore the concept of Multi-Trigger Backdoor Attacks (MTBAs), where multiple adversaries leverage different types of triggers to poison the same dataset. By proposing and investigating three types of multi-trigger attacks including \textit{parallel}, \textit{sequential}, and \textit{hybrid} attacks, we demonstrate that 1) multiple triggers can coexist, overwrite, or cross-activate one another, and 2) MTBAs easily break the prevalent shortcut assumption underlying most existing backdoor detection/removal methods, rendering them ineffective. Given the security risk posed by MTBAs, we have created a multi-trigger backdoor poisoning dataset to facilitate future research on detecting and mitigating these attacks, and we also discuss potential defense strategies against MTBAs. Our code is available at \url{https://github.com/bboylyg/Multi-Trigger-Backdoor-Attacks}.
\end{abstract}


\section{Introduction}

\label{sec:intro}
Deep neural networks (DNNs) have become the standard models for tasks in computer vision~\cite{he2016deep, dosovitskiy2020image}, natural language processing~\cite{devlin2018bert, mann2020language}, and speech recognition~\cite{chan2016listen}. However, research has demonstrated that DNNs are susceptible to backdoor attacks~\cite{gu2017badnets, chen2017targeted, li2020rethinking}, where stealthy triggers are embedded into the target models during training by poisoning a small subset of the training data or altering the training process. In image classification, this typically involves adding a fixed, carefully crafted trigger pattern to a few training images that the adversary can access. The objective of a backdoor attack is to manipulate the model into producing a specific output chosen by the adversary whenever the trigger pattern appears in a test input. With the increasing use of large models pre-trained on unsupervised web data or those provided by untrusted sources, concerns about the backdoor vulnerability of these models have grown, particularly when they are used in safety-critical applications.

\begin{figure}[!tp]
\small
\centering
\centerline{\includegraphics[width = .9\linewidth]{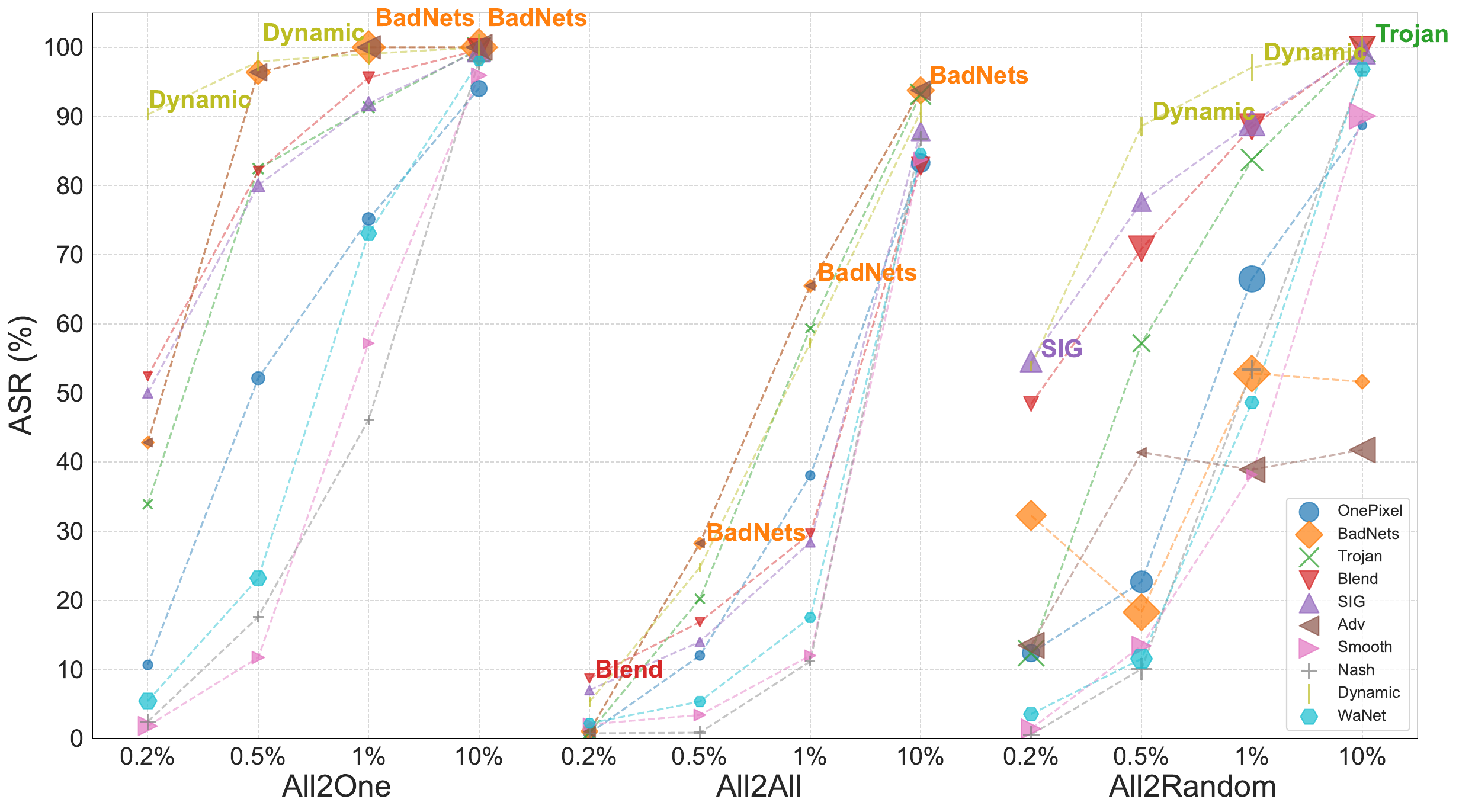}}
\caption{Effectiveness of multi-trigger attacks at various poisoning rates ($0.2\% \sim 10\%$) under 3 labeling modes (All2One, All2All, and All2Random) on the CIFAR-10 dataset. The results of All2One and All2All show that 1) different triggers can largely coexist at $10\%$ poisoning rate with high attack success rates (ASRs) but exhibit varied ASRs at extremely low poisoning rate ($0.2\%$). 
}
\vskip -0.15in
\label{fig:motivation}
\end{figure}

Given the significant security threat posed by backdoor attacks on neural networks, a variety of backdoor detection and removal techniques have been developed. Some prominent methods include~\cite{wang2019neural,li2021neural,wu2021adversarial,li2023reconstructive}. These methods generally operate under the assumption that backdoor triggers are unknown in practice. As a result, they focus on identifying backdoors by detecting the presence of ``shortcuts" between a specific source class and a target class, and then mitigating the threat by eliminating these shortcuts. However, in this work, we systematically demonstrate that such approaches can be effectively bypassed by implementing multiple backdoor triggers that create shortcuts in all directions, rendering them indistinguishable and making it challenging to target any specific one for removal. This strategy exposes a critical vulnerability in existing backdoor defenses, highlighting the need for more robust and comprehensive protection mechanisms.

In this work, we extend traditional single-trigger backdoor attacks to multi-trigger backdoor attacks. Note that we do not call them \emph{multi-adversary attacks} as our focus is on the triggers, demonstrating that even a single adversary can simultaneously employ multiple distinct triggers to poison a dataset. We explore three distinct implementation strategies for multi-trigger backdoor attacks: \emph{parallel}, \emph{sequential}, and \emph{hybrid}, where the latter blends multiple trigger patterns into a single, more potent super trigger pattern. Furthermore, we conduct a thorough analysis of 10 different existing triggers, uncovering their \emph{coexistence}, \emph{overwriting}, and \emph{cross-activation} effects. Our findings reveal that: 1) different backdoor triggers can coexist within the same dataset, whether injected in parallel or sequentially; 2) certain triggers can overwrite others even at minimal poisoning rates (as low as 0.2\% or 0.5\%); 3) one trigger can cross-activate another in sequential attack settings; and 4) different trigger patterns can be combined into a hybrid trigger to enhance attack efficacy. Fig. \ref{fig:motivation} summarizes some of the experimental results. These insights highlight the heightened security risks posed by the presence of multiple backdoor adversaries.

\begin{figure}[!tp]
\small
\centering
\centerline{\includegraphics[width = 0.85\linewidth]{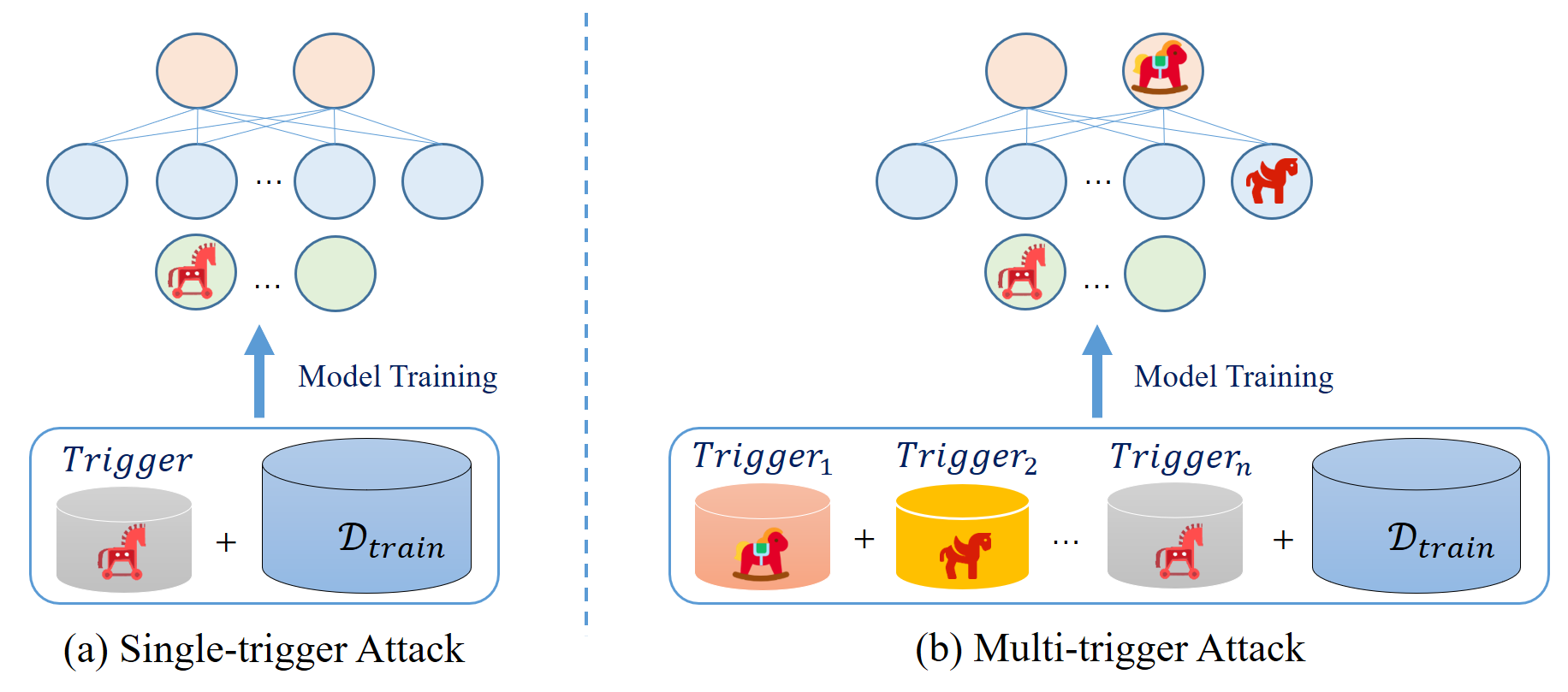}}
\caption{An illustrative comparison between single-trigger and multi-trigger backdoor attacks.
}
\vskip -0.15in
\label{fig:overview}
\end{figure}

Next, we systematically evaluate existing backdoor defense methods since they have not been adequately evaluated against multi-trigger attacks. This gap in evaluation can lead to overly optimistic perceptions of backdoor robustness and a false sense of security. Our findings confirm this vulnerability: multi-trigger backdoor attacks can severely undermine the effectiveness of current state-of-the-art backdoor detection and removal techniques. Specifically, 1) widely used detection methods like Neural Cleanse (NC) \cite{wang2019neural} struggle to identify multi-trigger attacks, and 2) mainstream backdoor removal techniques such as fine-tuning \cite{liu2018fine}, Neural Attention Distillation (NAD) \cite{li2021neural}, Adversarial Neuron Pruning (ANP) \cite{wu2021adversarial}, Reconstructive Neuron Pruning (RNP) \cite{li2023reconstructive}, and Anti-Backdoor Model (ABM) \cite{chen2017targeted} fail to effectively purify models compromised by multiple triggers. The challenge of identifying multiple triggers (or the correct sequence to defend against them) is compounded by the difficulty of determining the exact number of triggers present, making defense against multi-trigger attacks extremely difficult.

Our main contributions are as follows:

\begin{itemize}
\item We introduce the concept of multi-trigger backdoor attacks to expose the limitations of existing ``shortcut"-based backdoor detection and removal methods. We highlight the practical threat posed by such attacks in real-world scenarios, where a dataset or model could be simultaneously attacked by multiple adversaries using parallel, sequential, or hybrid triggers.

\item Through more than 200 experiments involving 10 types of triggers, 3 poisoning strategies (parallel, sequential, and hybrid-trigger), 4 poisoning rates (ranging from 0.2\% to 10\%), 3 label modification strategies (All2One, All2All, and All2Random), 2 datasets (CIFAR-10 and an ImageNet subset), and 4 DNN architectures (including 2 CNNs and 2 ViTs), we uncover the coexistence, overwriting, and cross-activation effects of multi-trigger backdoor attacks.

\item By re-evaluating 8 existing backdoor defense methods (4 for detection and 4 for removal) against multi-trigger backdoor attacks with 10 types of triggers, we demonstrate that: 1) all detection methods struggle with All2All and All2Random attacks, though they show some effectiveness against All2One attacks; 2) none of the detection methods can accurately identify the target of multi-trigger attacks; and 3) no backdoor removal method is capable of fully eliminating any of the 10 triggers.

\item We have built and released a multi-trigger backdoor dataset to support future research on backdoor attacks and defenses.

\end{itemize}

It is essential to emphasize that our work represents a significant attempt at designing MTBAs to simulate a more realistic scenario. The introduction of MTBAs itself undoubtedly provides a novel attack paradigm, offering new perspectives and design strategies for future research. We believe that MTBAs are an important leap compared to the existing single-trigger setting, and hope more researches could focus on potential threat of MTBAs. 

The remainder of this paper is organized as follows: In section II, we briefly review the related work on backdoor attacks and defenses. Following that, we present the threat model and formally define the problem studied in this paper. In Section IV, we introduce the technical details of our proposed multi-trigger backdoor attack strategies. Section V presents extensive experiments on multiple benchmark datasets, validating the effectiveness and impact of our method. In Section VI, we re-evaluate the performance of existing backdoor defense techniques against multi-trigger backdoor attacks and analyze the challenges they face. Finally, Section VII concludes the paper by summarizing our findings and discussing future research directions. We hope our work provides new insights into backdoor threats and motivates further research on more robust defense mechanisms.

\begin{table*}[t]
\centering
\caption{Comparison of existing multi-trigger backdoor attack studies and our MTBAs framework.}
\label{tab:mtba_comparison}
\renewcommand{\arraystretch}{1.2}
\setlength{\tabcolsep}{4pt}
\begin{adjustbox}{width=0.96\linewidth}
\begin{tabular}{l p{2.5cm} p{2.8cm} p{2.8cm} p{2.5cm} p{2.5cm} p{2.5cm}}
\toprule
\textbf{Paper \& Year} & \textbf{Trigger Type(s)} & \textbf{Poisoning Mode} & \makecell[c]{\textbf{Evaluation}\\\textbf{Scope}} & \makecell[c]{\textbf{Model}\\\textbf{Architecture}} & \makecell[c]{\textbf{Defense}\\\textbf{Evaluation}} & \makecell[c]{\textbf{Interaction}\\\textbf{Analysis}} \\
\midrule
Xue et al., TDSC 2022 \cite{xue2020one} & Static triggers (1–2) & Dataset-level, single shot & $\leq$2 triggers, 1 dataset & CNN only (CIFAR-10) & Basic (All2One) & Partial overwrite \\
Xue et al., Appl Intell 2024 \cite{xue2024imperceptible} & Imperceptible multi-channel & Sample-wise, single shot & $\leq$2 triggers, 1 dataset & CNN only & No & None \\
Singh et al., AISP 2023 \cite{singh2023physical} & Physical triggers, same target & Real-world setup, single shot & Physical attacks, 1 at a time & CNN only & No & None \\
Hou et al., TCSVT 2024 \cite{hou2024m} & Multi-trigger, multi-target & Single shot & $\leq$2 triggers, 1 dataset & CNN only & Limited & None \\
Sun et al., ICSIP 2024 \cite{sun2024invisible} & Frequency-domain, invisible & Multi-trigger, single sample & $\leq$2 triggers, 1 dataset & CNN only & No & None \\
\rowcolor{gray!15}
\textbf{MTBA (Ours)} & \textbf{10 diverse (static + sample-wise)} & \textbf{Parallel / Sequential / Hybrid} & \textbf{10 triggers $\times$ 2 datasets (CIFAR-10, ImageNet)} & \textbf{CNN + ViT (4 models)} & \textbf{Yes (8 defenses)} & \textbf{Yes (Coexistence, Overwrite, Cross-activation)} \\
\bottomrule
\end{tabular}
\end{adjustbox}
\end{table*}
\section{Related Work}

\subsection{Backdoor Attacks}
A backdoor attack implants a trigger pattern into a model so that the model predicts an adversary-chosen label when the trigger is present, while maintaining normal performance on clean data. Existing backdoor triggers can be broadly categorized into \emph{dataset-wise} and \emph{sample-wise} triggers. A dataset-wise trigger applies the same perturbation to all poisoned samples, such as a single pixel~\cite{tran2018spectral}, a checkerboard pattern~\cite{gu2017badnets}, global Gaussian noise~\cite{chen2017targeted}, or background reflection~\cite{liu2020reflection}. In contrast, sample-wise triggers generate diverse perturbations for each sample, often through optimization or generative models, e.g., composite backdoors~\cite{lin2020composite}, input-aware triggers~\cite{nguyen2020input}, deep generative triggers~\cite{cheng2021deep}, or invisible triggers~\cite{li2021invisible}.

Regardless of their diversity, most prior works adopt a \emph{single-trigger attack setting}, where there exists only one adversary and one type of trigger~\cite{li2022backdoor}. This restricted setting limits the understanding of how backdoors behave in realistic environments, where multiple adversaries may poison the same dataset with different triggers. This naturally raises the question of whether such attacks can coexist, compete, or interact when injected simultaneously.

Recent studies have explored \emph{multi-trigger attacks}. For example, Xue et al.\cite{xue2020one}, Singh et al.\cite{singh2023physical}, Hou et al.\cite{hou2024m}, and Sun et al.\cite{sun2024invisible} investigate scenarios with more than one trigger. However, these works remain limited—typically considering only two or three handcrafted triggers, restricted to CNN architectures, or focusing on narrow evaluation settings (e.g., All-to-One attacks on images).

\textbf{Comparison with our MTBAs.} Table~\ref{tab:mtba_comparison} contrasts existing multi-trigger backdoor attacks with our proposed MTBAs. The contribution of MTBAs extends beyond simply introducing multiple triggers:
\textit{1) Trigger types.} Prior work typically considers only a few handcrafted triggers (e.g., 1–2 visual patches or frequency-domain perturbations). In contrast, our benchmark incorporates 10 diverse triggers, including both static and sample-wise variants;
\textit{2) Attack modes.} Earlier studies largely rely on one-shot settings. Our framework introduces three poisoning strategies—parallel, sequential, and hybrid—that more realistically simulate coexistence or collusion between adversaries;
\textit{3) Evaluation scope.} Most prior works tested only a few triggers on a single CNN model. We systematically evaluate 10 triggers $\times$ 4 models (2 CNNs and 2 ViTs) $\times$ 2 datasets (CIFAR-10 and an ImageNet subset), providing a much broader and more rigorous scope;
\textit{4) Model architectures.} Previous studies restrict evaluations to CNNs, while our work includes both CNNs and Vision Transformers (ViTs), highlighting vulnerabilities across architectures;
\textit{5) Defense evaluation.} Earlier multi-trigger studies rarely evaluated defenses systematically. We assess eight mainstream defenses (four detection and four removal) under All-to-One, All-to-All, and All-to-Random settings, revealing their limitations;
\textit{6) Interaction analysis.} To our knowledge, no prior study investigates how triggers interact. Our work is the first to analyze \textit{coexistence, overwriting, and cross-activation effects}, showing that novel multi-trigger attack threats for evaluating real-world risks.

\subsection{Backdoor Defense} 
Existing backdoor defenses can be categorized into backdoor detection and backdoor removal (or mitigation) methods. Detection methods identify whether a given model has been backdoored by a backdoor attack \cite{chen2018detecting, wang2019neural, guo2019tabor,xu2021detecting} or whether a sample contains a backdoor trigger \cite{tran2018spectral,chen2018detecting,gao2019strip,zeng2021rethinking,huang2023distilling}.

Removal methods aim to eliminate the backdoor trigger (if it exists) from the model without affecting its functionality. This can be done during the training process via anti-backdoor learning \cite{li2021anti,huang2022backdoor}, or later on via fine-tuning, fine-pruning \cite{liu2018fine}, distillation \cite{li2021neural}, adversarial pruning \cite{wu2021adversarial}, channel Lipschitzness based pruning \cite{zheng2022data}, or reconstructive pruning \cite{li2023reconstructive}. The Anti-Backdoor Model (ABM) \cite{chen2024anti} is a representative example, exploiting trigger interaction properties to develop a non-invasive defense mechanism.
However, all of these defenses were developed for single-trigger attacks, thus facing high uncertainty when applied to defend multi-trigger attacks. In this paper, we run extensive experiments to reexamine the effectiveness of these defenses to multi-trigger backdoor attacks.

\section{Multi-Trigger Backdoor Attacks}
We first introduce our threat model and definition of backdoor attacks and then introduce the three types of multi-trigger backdoor attacks proposed in this work.


\subsection{Threat Model}
Our threat model introduces 3 poisoning strategies for multi-trigger attacks: \emph{\textbf{parallel}}, \emph{\textbf{sequential}}, and \emph{\textbf{hybrid}}, covering scenarios involving both independent and collusive adversaries: 

\begin{itemize}
    \item \textit{Parallel Attack:} In this scenario, independent adversaries act simultaneously but target different subsets of the training data. Each adversary introduces a distinct trigger or a set of triggers into their respective data portion. 

    \item \textit{Sequential Attacks:} In this strategy, adversaries poison the same subset of training samples in a sequential order. Each adversary introduces one or more triggers into these samples, stacking them one after another.
    \item \textit{Hybrid Attacks:} In the hybrid attack scenario, adversaries operate in a collusive manner, injecting multiple backdoor triggers into overlapping datasets, leading to a situation where the model is simultaneously influenced by a combination of parallel and sequential triggers. 
\end{itemize}

Specifically, in parallel and sequential MTBA, we validate the simultaneous existence of multiple independent attackers (up to 10) and analyze the case of independent attacks (as detailed in Sections \ref{sec:parallel_attack} and \ref{sec:sequential_attack}). In hybrid MTBA, we consider scenarios where multiple attackers collaborate to attack the same data samples (as detailed in Section \ref{sec:hybrid_attack}). This diversity allows us to gain a comprehensive understanding of how different types of triggers can coexist and interfere with each other, which can inspire backdoor research in the future.

\subsection{Definition of Backdoor Attack}
We focus on image classification tasks. Given a clean training dataset $\gD = \{(\vx_i, y_i)\}_{i=1}^{N}$, where $\vx_i \in \gX$ represents a training image and $y_i \in \gY$ is its label. A backdoor adversary generates backdoor examples with a triggering function $tr: \gX \rightarrow \gX$. For each clean sample, it maps the sample $\vx$ into a backdoor sample $\vx_b$, i.e., $\vx_b= \tr(\vx)$ and modifies its label to a backdoor target label $y_{b}$.  To ensure stealthiness and efficacy for backdoor injection, the adversary randomly chooses a few training samples to poison, which creates a set of backdoor samples $\gD_{b} = \{(\vx_b, y_{b})\}$. As we mentioned earlier, most existing attacks only inject a single type of backdoor trigger into the training data, meaning there is only one triggering function $\tr(\cdot)$. We call these attacks \emph{Single-Trigger Backdoor Attacks (STBAs)}. The final poisoned training dataset can be denoted as $\hat{\gD} = \gD_{c} \cup \gD_{b}$, where $\gD_{c} = \{(\vx_c, y_c)\}$ represents clean samples and clean labels, while $\gD_{b} = \{(\vx_b, y_b)\}$ represents backdoor samples and backdoor labels. The poisoning rate is defined as $\alpha = |\gD_{b}|/|\hat{\gD}|$. Training a model on $\hat{\gD}$ is solving the following optimization problem:
\begin{equation}\label{eq:single-trigger}
\min_\theta \, \E_{\gD_{c}} [\Ls(f_{\theta}(\vx_c), y_c )] + \E_{\gD_{b}} [\Ls(f_{\theta}(\vx_b),y_b)],
\end{equation}
where $\Ls$ is the cross-entropy loss. The first term in the above objective formulates the loss of the clean (original) task, while the second term formulates the loss of the backdoor task.
Training on the dataset can be viewed as a process where the model learns both tasks.

\subsection{Definition of Multi-Trigger Backdoor Attack}
We call backdoor attacks that utilize multiple types of triggers (possibly from one or more adversaries) to attack the same dataset termed \emph{Multi-Trigger Backdoor Attacks} (MTBAs). Fig. \ref{fig:overview} illustrates the idea of MTBA. Note that we did not call these attacks multi-adversary attacks as even one adversary could implement different types of triggers. The MTBA problem can also be formulated as \eqref{eq:single-trigger}, but with a slightly more complex poisoning set that contains multiple triggers:
\begin{equation}\label{eq:multi-trigger}
    \gD_b= \bigcup\limits_{k=1}^{m} \gD^k_{b}=\bigcup\limits_{k=1}^{m} \{(\vx^k_b, y^k_b)\},
\end{equation}
where $\vx^k_b = \tr_k(\vx)$ is a poisoned sample by the $k$-th trigger $\tr_k(\cdot)$, and $\gD^k_{b}$ is the subset of poisoned samples by trigger $\tr_k(\cdot)$ for overall $m$ triggers. For multi-trigger attacks, the poisoned dataset becomes $\hat{\gD}=\gD_c \cup \gD_b$ with a poisoning rate of $\alpha = |\gD_{b}|/|\hat{\gD}| = \sum_{k=1}^{m}|\gD^k_{b}|/|\hat{\gD}|$. 

\section{Poisoning Strategies of MTBAs}\label{sec:poisoning}
Conceptually, MTBA is an extension of STBA. While STBA injects only one type of trigger into the target model, MTBA introduces multiple types of triggers.  We introduce three poisoning strategies to simulate diverse real-world threat scenarios: \emph{parallel}, \emph{sequential}, and \emph{hybrid-trigger}.

\subsection{Parallel Poisoning} \label{sec:parallel_attack}
Arguably, the triggers can be injected into the victim dataset by multiple independent adversaries. In this case, it is reasonable to assume that the poisoned subsets are not overlapping with each other, as two independent adversaries are of extremely low probability to poison the same sample, given the low poisoning rate. We call this poisoning strategy \emph{parallel poisoning}. 

To implement parallel poisoning, we randomly sample a few training samples into a backdoor candidate subset $\gD_s$ and then uniformly divide $\gD_s$ into $m$ smaller subsets, i.e., $\gD_s = \{\gD_s^{k}\}_{k=1}^{m}$. We then assign, to each $\gD_s^{k}$, a randomly selected trigger $\tr_{k} \in \gT = \{\tr_k\}_{k=1}^{m}$ from a trigger pool, i.e., $\gD_{b}^{k} = \{(\vx_b^k,y_b^k)|\vx_b^k=\tr_{k}(\vx_s^k), \vx_s^k \in \gD_{s}^{k})\}$. Accordingly, training a model on multi-trigger poisoned dataset $\{\gD_{b}^{k}\}_{k=1}^{m}$ can be formulated as:
\begin{equation}
\min_\theta \, \E_{\sim \gD_c} [\Ls(f_\theta(\vx_c), y_c)] + \sum_{k=1}^{m} \E_{\gD_b^{k}} [\Ls(f_\theta(\vx_b^k), y_b^k)],
\end{equation}
where $\Ls$ is the cross-entropy loss and $m$ is the number of independently poisoned subsets. Training on a multi-trigger backdoored dataset can be viewed as the learning process of one clean task and $m$ backdoor tasks simultaneously.

\subsection{Sequential Poisoning} \label{sec:sequential_attack}
It is also possible that different adversaries launch their attacks at different times. In this case, different adversaries may attack the same dataset in sequential order, but still on non-overlapping data subsets. We call this attacking strategy as \emph{sequential poisoning}. This poisoning strategy allows us to study the overwriting effect of different triggers, i.e., the question of whether an early-injected trigger can stay effective in the presence of subsequent attacks.

To implement sequential poisoning, we inject different types of triggers into the victim dataset following a specific order. Suppose adversary $k$ can poison a small subset of clean samples $\gD_s^k$ with its own trigger $\tr_k$ to obtain a backdoor subset $\gD_b^k$, and accordingly a poisoned training dataset $\hat{\gD}_k=\gD_c\cup\gD_b^{k}=\{\gD\setminus\gD_s^k\}\cup\gD_b^k$, where $\gD$ is the original dataset, $\gD_c$ is the subset of remaining clean samples, $\gD_s^k$ is the victim clean subset, $\gD_b^k$ are backdoor samples generated from $\gD_s^k$. The model is then trained on the poisoned dataset to obtain a backdoored model $f_{\theta_k}$, as follows:
\begin{equation}\label{eq:k}
    \min_{\theta_k} \, \E_{(\vx,y) \sim \hat{\gD}_k} [\Ls(f_{\theta_k}(\vx), y)].
\end{equation}
The next adversary $k+1$ poisoned the dataset $\hat{\gD}_k$ following the same procedure as adversary $k$ to obtain poisoned dataset $\hat{\gD}_{k+1}=\gD_c\cup\gD_b^{k}\cup\gD_b^{k+1}=\{\hat{\gD}_k\setminus\gD_s^{k+1}\}\cup\gD_b^{k+1}$. The backdoored model $f_{\theta_k}$ is then continuously trained on $\hat{\gD}_{k+1}$ to produce  $f_{\theta_{k+1}}$, as follows:
\begin{equation}\label{eq:k+1}
    \min_{\theta_{k+1}} \, \E_{(\vx,y) \sim \hat{\gD}_{k+1}} [\Ls(f_{\theta_{k+1}}(\vx), y)].
\end{equation}
After the above sequential training, model $f_{\theta_{k+1}}$ becomes a sequentially backdoored model that contains both trigger $\tr_k$ and trigger $\tr_{k+1}$.

Note that, although the adversaries follow a sequential order, we assume they are independent and their selected victim subsets do not overlap ($\gD_s^{k} \cap \gD_s^{k+1}=\emptyset$) and only contain clean samples. These are reasonable assumptions because: \textbf{1)} the adversaries have access to the victim samples, thus can easily ensure that they are clean (so would not impact trigger injection); and \textbf{2)} independent attackers often own different samples. We also assume the subsequent adversary trains the backdoored model on the basis of the previous backdoored model. This is to simulate the current trend of fine-tuning large models, where a victim user may download and fine-tune a pre-trained (and poisoned by a previous adversary) large model for its own downstream application. 

\subsection{Hybrid-trigger Poisoning} \label{sec:hybrid_attack}
There may also exist a super adversary that combines different triggers into one hybrid trigger to achieve the effect of multiple triggers. This poisoning strategy is realistic because the current literature already has a large number of trigger patterns for the adversary to exploit. We call this poisoning strategy as  \emph{hybrid-trigger poisoning}.

\begin{figure}[!tp]
\small
\centering
\centerline{\includegraphics[width = 1.0\linewidth]{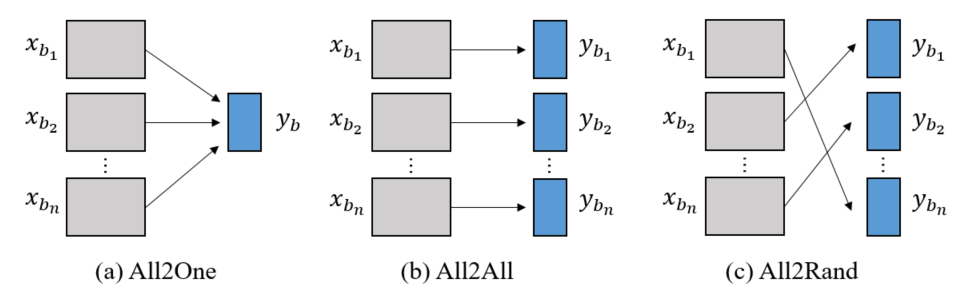}}
\caption{An illustration of the three label modification strategies.
}
\vskip -0.15in
\label{fig:label modification}
\end{figure}

In contrast to parallel and sequential poisonings, which are based on multiple independent triggers, hybrid-trigger poisoning represents a sample-wise attacking strategy. It simultaneously introduces multiple distinct triggers into one single input sample, endowing it with multi-trigger characteristics. Specifically, given a clean sample $\vx$, a hybrid-trigger attack poisons the sample with $m$ elementary triggers $\gT = \{\tr_k\}_{k=1}^{m}$ as follows:
\begin{equation}\label{eq:hybrid-trigger}
    \vx_h = \tr_m\circ\tr_{m-1}\circ\cdots\circ\tr_1(\vx),
\end{equation}
where, we use soft blending at each step, i.e., $\tr_k\circ\tr_{k-1}(\vx)= \lambda\cdot\tr_k + (1-\lambda) \vx_b^{k-1}$ (we set $\lambda=0.25$ in our experiments).
Suppose the training dataset is $\gD$ and the small subset of clean samples accessible to the adversary is $\gD_s \subset \gD$, the adversary injects the hybrid trigger into $\gD_s$ to obtain the backdoor subset $\gD_h=\{(\vx_h,y_h)\}$ following \eqref{eq:hybrid-trigger}. The poisoned dataset can then be defined as $\hat{\gD} = \gD_c\cup\gD_h=\{\gD\setminus\gD_s\}\cup\gD_h$. The adversary can then train a backdoored model on $\hat{\gD}$ following:
\begin{equation}\label{eq:hybrid-attack}
\min_\theta \, \E_{\gD_{c}} [\Ls(f_{\theta}(\vx_c), y_c )] + \E_{\gD_{h}} [\Ls(f_{\theta}(\vx_h),y_h)].
\end{equation}

\noindent\textbf{Candidate Triggers.} Gathering a representative and meaningful set of triggers $\gT$ is crucial for our study. By investigating the current literature, we meticulously selected 10 types of triggers used in mainstream backdoor attacks as our candidate triggers, which include both dataset-wise triggers (such as one pixel, a checkerboard pattern, and adversarial perturbation) and sample-wise triggers (such as input-aware pattern and image deformation). A detailed list of the studied triggers can be found in Section \ref{sec:experimental setting}.

\subsection{Label Modification}
In the context of multi-trigger backdoor attacks, different adversaries may share the same backdoor target if it is of general interest, or they may have entirely different (or random) target labels. Following prior works \cite{gu2017badnets,xue2020one}, we explore the following three label modification strategies, including \emph{All2One}, \emph{All2All}, and \emph{All2Random} (see Fig. \ref{fig:label modification}).

\begin{itemize}
\item \textbf{All2One:} This strategy relabels all backdoor samples to a fixed backdoor target label $y_t$, i.e., all samples $\vx_{b/h} \in \gD_{b/h}$ have the same label $y_t$. In other words, all adversaries share the same backdoor target.

\item \textbf{All2All:} It modifies the label of a backdoor sample $\vx_b$ (crafted from clean sample $\vx$) to $y_{b}= (y+1) \% K$, where $K$ is the total number of classes, $y$ is the original (clean) label of $\vx/\vx_b$ and $y_{b}$ is its modified label. This is to simulate the scenario where the next class is of particular interest to all adversaries.

\item \textbf{All2Random:} This strategy simulates the scenarios where there exists no common target between the adversaries and each label has an equal chance to be selected as the target label. In this case, we modify the label of a backdoor sample $\vx_b$ to $y_b = Random(\{1,2,\cdots, K\})$ where $Random(\cdot)$ is a random function.
\end{itemize}

\section{Experiments}
In this section, we experiment and summarize the set of key findings obtained with multi-trigger attacks, including the \textit{coexisting, overwriting, and cross-activating} effects of different triggers under parallel, sequential, and hybrid-trigger settings, and the reliability of existing defenses in the presence of multi-trigger attacks.

\subsection{Experimental Setting}\label{sec:experimental setting}
\noindent\textbf{Attacks Setup.} We choose 10 representative triggers from the current literature as our candidate triggers, which include \emph{static triggers} like OnePixel \cite{tran2018spectral}, BadNets \cite{gu2017badnets}, Trojan attack \cite{liu2018trojaning}, \emph{global triggers} like Blend \cite{chen2017targeted}, Sinusoidal Spectrum (SIG) \cite{barni2019new}, adversarial noise (Adv) \cite{turner2019clean}, Smooth \cite{zeng2021rethinking}, Nashivell filter (Nash) \cite{liu2019abs}, and \emph{sample-wise triggers} like Dynamic \cite{nguyen2020input}, WaNet \cite{nguyen2021wanet}. To ensure a consistent and fair comparison, we employ a dirty-label poisoning setup for all triggers, which includes data poisoning and label modification in two steps. We test all triggers for the parallel, sequential, and hybrid-trigger poisoning strategies and All2One, All2All, and All2Random label modification strategies.  

We trained the backdoored models of the CNN architecture from scratch for 60 epochs, using SGD optimizer, batch size 128, weight decay $5\times10^{-4}$, and an initial learning rate of 0.1 which was decayed to 0.01 at the 40th epoch. For the Transformer architecture, we fine-tuned the model with publicly available pre-trained weights\footnote{https://github.com/huggingface/pytorch-image-models} for 5 to 10 epochs, while injecting the backdoor with a fine-tuning (learning) rate of 0.001 (other parameters were kept unchanged). All backdoored models were trained with standard data augmentation techniques including random crop and horizontal flipping. \\

\noindent\textbf{Models and Datasets.} For the target model, we mainly focus on two classical CNN architectures, including ResNet \cite{he2016deep} and MobileNet \cite{howard2017mobilenets}, and two Transformer architectures including ViT-base \cite{dosovitskiy2020image} and ViT-small. The four architectures are the most widely adopted models in standard, resource-limited, or large-scale computer vision applications. Moreover, there is little study of backdoor attacks and defenses with ViTs. We fill this gap through comprehensive studies with two ViT architectures. 
For the datasets, we consider two commonly adopted image datasets in the field: CIFAR-10 and an ImageNet \cite{deng2009imagenet} subset (the first 20 classes). Unless otherwise stated, we set the poisoning rate to 10\% (1\% for each type of trigger). The datasets and DNN models used in our experiments are summarized in Table \ref{tab:datasets}. The generated multi-trigger datasets will be released to help the development of more advanced backdoor defenses. Also, the poisoned datasets can continue to incorporate more advanced and future triggers following our findings. \\

\begin{table}[tp]
\renewcommand{\arraystretch}{1.35}
\centering
\caption{Datasets and models used in our experiments.}
\begin{adjustbox}{width=0.98\linewidth}
\begin{tabular}{ccccc}
\toprule
\textbf{Dataset} & \textbf{Model} & \textbf{No. Classes} & \textbf{Input Size} & \textbf{No. Training Images} \\ \hline
\multirow{4}{*}{CIFAR10} & ResNet-18 & 10 & $32\times32\times3$ & 50000 \\
 & MobileNet-V2 & 10 & $32\times32\times3$ & 50000 \\
 & ViT-Small & 10 & $224\times224\times3$ & 50000 \\
 & ViT-Base & 10 & $224\times224\times3$ & 50000 \\ \hline
\multirow{3}{*}{ImageNet-20 Subset} & ResNet-50 & 20 & $224\times224\times3$ &  26000\\
 & ViT-Base & 20 & $224\times224\times3$ &  26000\\ \bottomrule
\end{tabular}\label{tab:datasets}
\end{adjustbox}
\end{table}

\begin{figure}[tp]
\small
\centering
\centerline{\includegraphics[width = .9\linewidth]{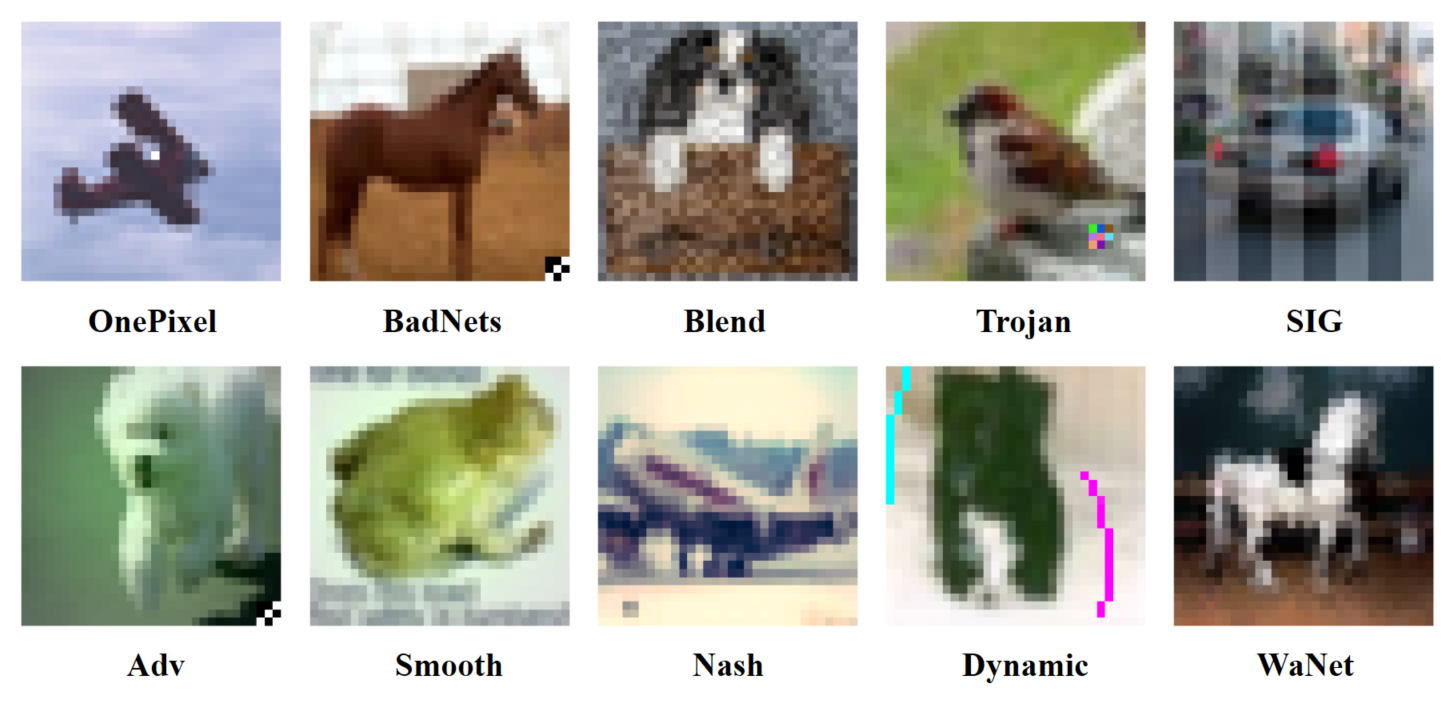}}
\caption{Examples of 10 types of backdoor triggers.
}
\vskip -0.15in
\label{fig:samples}
\end{figure}

\begin{table*}[!tp]
\renewcommand{\arraystretch}{1.25} 
\centering
\caption{The ASR (\%) of \emph{parallel} MTBAs on CIFAR-10, averaged over the 4 model architectures (i.e. ResNet-18, MobileNet-V2, VIT-Small, and VIT-Base). '10\% (1\%)' represents the total (trigger-wise) poisoning rate. The best attack performances are \textbf{boldfaced}. }
\begin{adjustbox}{width=0.94\linewidth}
\begin{tabular}{c|c|ccccccccccc|c}
\toprule
 &  & \multicolumn{12}{c}{\textbf{Parallel MTBAs (averaged over 4 model architectures)}} \\ \cline{3-14} 
\multirow{-2}{*}{\textbf{\begin{tabular}[c]{@{}c@{}}Label\\ Modification\end{tabular}}} & \multirow{-2}{*}{\textbf{\begin{tabular}[c]{@{}c@{}}Poisoning\\ Rate\end{tabular}}} & \multicolumn{1}{c|}{Clean Acc.} & OnePixel & BadNets & Trojan & Blend & SIG & Adv & Smooth & Nash & Dynamic & WaNet & \textbf{Average} \\ \hline
 & 10\% (1\%) & \multicolumn{1}{c|}{95.69} & 94.04 & 100.00 & 99.73 & 99.61 & 99.56 & 99.75 & 95.93 & 98.08 & 100.00 & 98.01 & 98.50 \\
 & 1\% (0.1\%) & \multicolumn{1}{c|}{95.70} & 75.16 & 100.00 & 91.33 & 95.59 & 91.85 & 99.25 & 57.18 & 46.12 & 99.05 & 73.05 & 82.93 \\
 & 0.5\% (0.05\%) & \multicolumn{1}{c|}{95.53} & 52.12 & 96.38 & 82.44 & 82.09 & 80.04 & 92.76 & 11.72 & 17.60 & 97.94 & 23.19 & 63.99 \\
 & 0.2\% (0.02\%) & \multicolumn{1}{c|}{95.94} & 10.64 & 42.83 & 33.89 & 52.38 & 49.98 & 38.40 & 1.81 & 2.43 & 90.28 & 5.43 & 33.25 \\ \cline{2-14} 
\multirow{-5}{*}{All2One} & \cellcolor[HTML]{C0C0C0}\textbf{Average} & \multicolumn{1}{c|}{\cellcolor[HTML]{C0C0C0}\textbf{95.71}} & \cellcolor[HTML]{C0C0C0}\textbf{57.99} & \cellcolor[HTML]{C0C0C0}\textbf{84.80} & \cellcolor[HTML]{C0C0C0}\textbf{76.85} & \cellcolor[HTML]{C0C0C0}\textbf{82.42} & \cellcolor[HTML]{C0C0C0}\textbf{80.36} & \cellcolor[HTML]{C0C0C0}\textbf{82.54} & \cellcolor[HTML]{C0C0C0}\textbf{41.66} & \cellcolor[HTML]{C0C0C0}\textbf{41.06} & \cellcolor[HTML]{C0C0C0}\textbf{96.82} & \cellcolor[HTML]{C0C0C0}\textbf{49.92} & \cellcolor[HTML]{C0C0C0}\textbf{69.67} \\ \hline
 & 10\% (1\%) & \multicolumn{1}{c|}{95.11} & 83.25 & 93.75 & 93.20 & 82.80 & 87.85 & 92.75 & 83.80 & 86.75 & 90.65 & 84.65 & 88.04 \\
 & 1\% (0.1\%) & \multicolumn{1}{c|}{95.62} & 38.05 & 65.50 & 59.35 & 29.65 & 28.40 & 64.39& 12.00 & 11.15 & 57.25 & 17.50 & 38.43 \\
 & 0.5\% (0.05\%) & \multicolumn{1}{c|}{95.73} & 12.00 & 28.25 & 20.20 & 16.85 & 14.00 & 27.13 & 3.40 & 0.85 & 24.75 & 5.35 & 15.39 \\
 & 0.2\% (0.02\%) & \multicolumn{1}{c|}{95.51} & 0.75 & 1.05 & 0.85 & 8.70 & 7.00 & 0.60 & 2.05 & 0.75 & 5.30 & 2.20 & 2.97 \\ \cline{2-14} 
\multirow{-5}{*}{All2All} & \cellcolor[HTML]{C0C0C0}\textbf{Average} & \multicolumn{1}{c|}{\cellcolor[HTML]{C0C0C0}\textbf{95.49}} & \cellcolor[HTML]{C0C0C0}\textbf{33.51} & \cellcolor[HTML]{C0C0C0}\textbf{47.14} & \cellcolor[HTML]{C0C0C0}\textbf{43.40} & \cellcolor[HTML]{C0C0C0}\textbf{34.50} & \cellcolor[HTML]{C0C0C0}\textbf{34.31} & \cellcolor[HTML]{C0C0C0}\textbf{46.22} & \cellcolor[HTML]{C0C0C0}\textbf{25.31} & \cellcolor[HTML]{C0C0C0}\textbf{24.87} & \cellcolor[HTML]{C0C0C0}\textbf{44.49} & \cellcolor[HTML]{C0C0C0}\textbf{27.42} & \cellcolor[HTML]{C0C0C0}\textbf{36.21} \\ \hline
 & 10\% (1\%) & \multicolumn{1}{c|}{95.30} & 88.74 & 51.61 & 99.73 & 99.73 & 99.45 & 41.76 & 90.05 & 96.39 & 99.71 & 96.80 & 86.40 \\
 & 1\% (0.1\%) & \multicolumn{1}{c|}{95.52} & 66.49 & 52.81 & 83.70 & 88.48 & 89.08 & 38.89 & 38.19 & 53.37 & 97.09 & 48.61 & 65.67 \\
 & 0.5\% (0.05\%) & \multicolumn{1}{c|}{95.50} & 22.67 & 18.26 & 57.18 & 70.83 & 77.63 & 41.38 & 13.42 & 10.06 & 88.59 & 11.54 & 41.16 \\
 & 0.2\% (0.02\%) & \multicolumn{1}{c|}{96.15} & 12.36 & 32.26 & 12.36 & 48.40 & 54.60 & 13.48 & 1.40 & 0.56 & 53.89 & 3.53 & 23.28 \\ \cline{2-14} 
\multirow{-5}{*}{All2Random} & \cellcolor[HTML]{C0C0C0}\textbf{Average} & \multicolumn{1}{c|}{\cellcolor[HTML]{C0C0C0}\textbf{95.62}} & \cellcolor[HTML]{C0C0C0}\textbf{47.57} & \cellcolor[HTML]{C0C0C0}\textbf{38.73} & \cellcolor[HTML]{C0C0C0}\textbf{63.24} & \cellcolor[HTML]{C0C0C0}\textbf{76.86} & \cellcolor[HTML]{C0C0C0}\textbf{80.19} & \cellcolor[HTML]{C0C0C0}\textbf{33.88} & \cellcolor[HTML]{C0C0C0}\textbf{35.77} & \cellcolor[HTML]{C0C0C0}\textbf{40.09} & \cellcolor[HTML]{C0C0C0}\textbf{84.82} & \cellcolor[HTML]{C0C0C0}\textbf{40.12} & \cellcolor[HTML]{C0C0C0}\textbf{54.13}  \\ \bottomrule
\end{tabular}
\end{adjustbox}
\label{tab:parallel_cifar}
\end{table*}

\begin{table*}[!tp]
\renewcommand{\arraystretch}{1.25} 
\centering
\caption{The ASR (\%) of parallel multi-trigger attacks on the ImageNet-20 subset. The best results are \textbf{boldfaced}.}
\begin{adjustbox}{width=0.94\linewidth}
\begin{tabular}{c|c|cccccc|cccccc}
\toprule
&  & \multicolumn{6}{c|}{\textbf{Parallel Attacks (Poisoning rate 10\%)}} & \multicolumn{6}{c}{\textbf{Parallel Attacks (Poisoning rate 1\%)}} \\ \cline{3-14} 
\multirow{-2}{*}{\textbf{\begin{tabular}[c]{@{}c@{}}Label\\ Modification\end{tabular}}} & \multirow{-2}{*}{\textbf{Model}} & Clean & BadNets & Trojan & Blend & SIG & Nash & Clean & BadNets & Trojan & Blend & SIG & Nash \\ \hline
 & ResNet-50 & 76.90 & 90.20 & 18.60 & 92.40 & 73.10 & 92.10 & 78.50 & 6.20 & 6.70 & 38.20 & 50.90 & 52.00 \\
 & ViT-Base & 93.10 & 97.90 & 43.90 & 99.70 & 96.90 & 95.80 & 93.70 & 9.30 & 5.90 & 91.70 & 55.20 & 72.20 \\
\multirow{-3}{*}{All2One} & \cellcolor[HTML]{C0C0C0}\textbf{Average} & \cellcolor[HTML]{C0C0C0}\textbf{85.00} & \cellcolor[HTML]{C0C0C0}\textbf{94.05} & \cellcolor[HTML]{C0C0C0}\textbf{31.25} & \cellcolor[HTML]{C0C0C0}\textbf{96.05} & \cellcolor[HTML]{C0C0C0}\textbf{85.00} & \cellcolor[HTML]{C0C0C0}\textbf{93.95} & \cellcolor[HTML]{C0C0C0}\textbf{86.10} & \cellcolor[HTML]{C0C0C0}\textbf{7.75} & \cellcolor[HTML]{C0C0C0}\textbf{6.30} & \cellcolor[HTML]{C0C0C0}\textbf{64.95} & \cellcolor[HTML]{C0C0C0}\textbf{53.05} & \cellcolor[HTML]{C0C0C0}\textbf{62.10} \\ \hline
 & ResNet-50 & 75.50 & 47.80 & 16.10 & 52.20 & 39.80 & 54.10 & 80.00 & 3.90 & 4.70 & 11.10 & 10.40 & 12.60 \\
 & ViT-Base & 93.70 & 91.60 & 38.50 & 89.70 & 86.10 & 86.90 & 94.10 & 24.50 & 2.50 & 41.30 & 22.40 & 34.90 \\
\multirow{-3}{*}{All2All} & \cellcolor[HTML]{C0C0C0}\textbf{Average} & \cellcolor[HTML]{C0C0C0}\textbf{84.60} & \cellcolor[HTML]{C0C0C0}\textbf{69.70} & \cellcolor[HTML]{C0C0C0}\textbf{27.30} & \cellcolor[HTML]{C0C0C0}\textbf{70.95} & \cellcolor[HTML]{C0C0C0}\textbf{62.95} & \cellcolor[HTML]{C0C0C0}\textbf{70.50} & \cellcolor[HTML]{C0C0C0}\textbf{87.05} & \cellcolor[HTML]{C0C0C0}\textbf{14.20} & \cellcolor[HTML]{C0C0C0}\textbf{3.60} & \cellcolor[HTML]{C0C0C0}\textbf{26.20} & \cellcolor[HTML]{C0C0C0}\textbf{16.40} & \cellcolor[HTML]{C0C0C0}\textbf{23.75} \\ \hline
 & ResNet-50 & 76.60 & 83.90 & 20.10 & 95.50 & 69.30 & 92.00 & 76.30 & 6.90 & 5.80 & 38.80 & 39.60 & 62.20 \\
 & ViT-Base & 93.60 & 99.00 & 46.60 & 99.80 & 96.10 & 97.70 & 93.80 & 70.30 & 5.60 & 94.00 & 72.40 & 64.20 \\
\multirow{-3}{*}{All2Random} & \cellcolor[HTML]{C0C0C0}\textbf{Average} & \cellcolor[HTML]{C0C0C0}\textbf{85.10} & \cellcolor[HTML]{C0C0C0}\textbf{91.45} & \cellcolor[HTML]{C0C0C0}\textbf{33.35} & \cellcolor[HTML]{C0C0C0}\textbf{97.65} & \cellcolor[HTML]{C0C0C0}\textbf{82.70} & \cellcolor[HTML]{C0C0C0}\textbf{94.85} & \cellcolor[HTML]{C0C0C0}\textbf{85.05} & \cellcolor[HTML]{C0C0C0}\textbf{38.60} & \cellcolor[HTML]{C0C0C0}\textbf{5.70} & \cellcolor[HTML]{C0C0C0}\textbf{66.40} & \cellcolor[HTML]{C0C0C0}\textbf{56.00} & \cellcolor[HTML]{C0C0C0}\textbf{63.20} \\
\bottomrule
\end{tabular}\label{tab:parallel_imagenet}
\end{adjustbox}
\end{table*}

\noindent\textbf{Defense Setup.} 
We considered 9 advanced backdoor defense methods, including 4 backdoored model detection techniques: Neural Cleanse (NC) \cite{wang2019neural}, UMD \cite{xiang2023umd}, MMDB \cite{wang2023mm}, and Unlearning \cite{li2023reconstructive}, as well as 4 mainstream backdoor removal techniques, including Fine-tuning (FT), Finepruning (FP) \cite{liu2018fine}, Neural Attention Distillation (NAD) \cite{li2021neural}, Adversarial Neural Pruning (ANP) \cite{wu2021adversarial} and Anti-backdoor Model (ABM) \cite{chen2024anti}. For all defense methods, we used their open-source code and carefully adjusted hyperparameters to achieve the best defense performance. All the above defense techniques used the same data augmentation techniques as model training, i.e., random crop and horizontal flipping.

\subsection{Evaluating and Understanding MTBAs}
We first evaluate MTBAs with parallel, sequential, and hybrid-trigger poisonings, respectively. With these experiments, we further reveal the coexisting, overwriting, and cross-activating effects between different triggers. \\

\subsubsection{Parallel MTBAs} \label{sec:parallel_mtba}
Recall that, in parallel MTBAs, we inject the 10 triggers all together into the target CIFAR-10 and ImageNet-20 datasets, with each trigger poisoning a unique subset of clean samples. Here, we report the attack performance on CIFAR-10 and ImageNet-20 subset. Table \ref{tab:parallel_cifar} summarizes the attack success rate of the 10 triggers under various poisoning rates $\alpha\in[0.2\%,10\%]$. 

\noindent\textbf{Coexisting Effect.}
We first look at the average ASR (last column) result at poisoning rate $10\%$. One key observation is that these triggers can coexist well in a single model, achieving 98.50\%, 88.04\%, and 86.40\% ASR for All2One, All2All, and All2Random targets, respectively. Most of the triggers demonstrate an ASR of well above 80\%, except for the BadNets and Adv triggers under the All2Random attacks. This indicates that fixed trigger patterns (Adv is an adversarially perturbed version of the BadNets checkerboard trigger) may be easily influenced by other triggers. Overall, All2One attacks exhibit a more robust attack performance than All2All or All2Random attacks at this poisoning rate, which is somewhat expected as All2One attacks have the same target.

Moving on to low poisoning rates (0.2\% to 1\%), we find that the average ASR over all 10 triggers (the last column) degrades significantly with the poisoning rate. However, there are still survivors at an extremely low poisoning rate of 0.2\%, for example, the Blend, SIG, and Dynamic triggers have an ASR $> 40\%$ under the All2One and All2Random settings. No triggers are successful under the All2All setting at this low poisoning rate. We conjecture this is because the cyclic relabeling strategy of All2All causes both local (intra-trigger) and global (inter-trigger) overlaps and disruptions to the ``trigger-target'' mapping, making it difficult to associate a small fixed trigger to rotated target labels.  This result indicates that the All2All setting turns out to be the most challenging attack setting under extremely low poisoning rates (e.g., 0.2\% and 0.5\%). 
This indicates that if the adversaries have overlapped targets but different triggers, they are less likely to succeed at low poisoning rates. However, we do not attribute this phenomenon to the overwriting effect between different triggers, but rather to the varying strengths of the triggers themselves, as (almost) all triggers fail badly.

The results in Table \ref{tab:parallel_cifar} can also help answer the question as to whether there exists a single strongest trigger in all scenarios. It shows that, under the All2One setting, Dynamic is the strongest trigger on average across different poisoning rates; under the All2All setting, the BadNets trigger is the strongest; while under the All2Random setting, the Dynamic trigger becomes the strongest again. Overall, the Dynamic trigger is the strongest under the All2One and All2Random settings and is close to the best trigger (BadNets) under the All2All setting. Therefore, we believe it has the potential to be the single strongest trigger. Our experiments with sequential MTBAs in Section \ref{sec:seqential_mtba} also confirm the hybrid characteristic of the Dynamic trigger, i.e., it can cross-activate other triggers. It is worth mentioning that not all sample-wise triggers are strong, e.g., sample-wise triggers Smooth and WaNet are weaker than dataset-wise trigger BadNets under the All2One and All2All settings.

\begin{figure}[tp]
\begin{minipage}[b]{0.32\linewidth}
\centering
\includegraphics[width=0.99\textwidth]{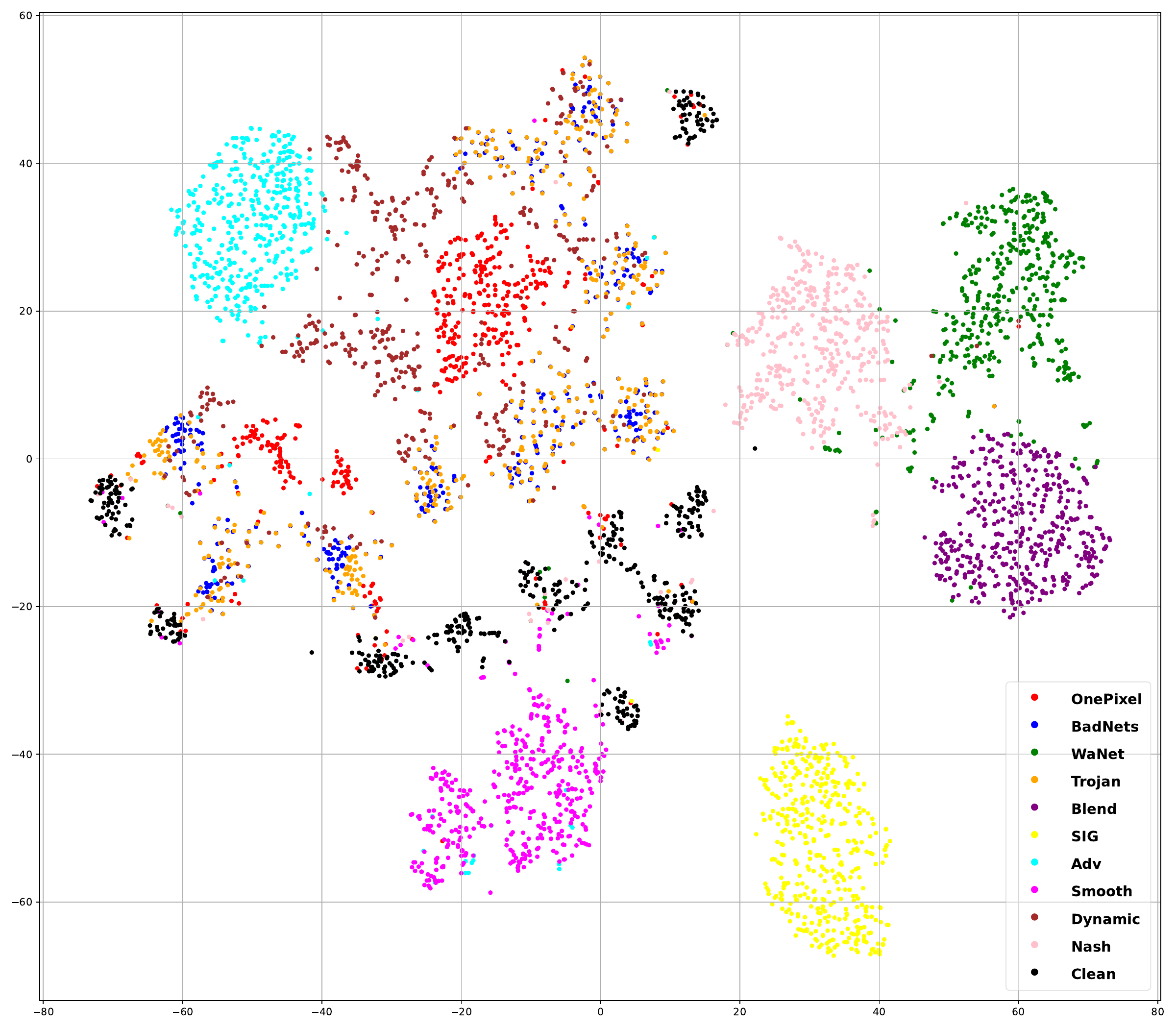}
\centerline{(a) All2One}
\end{minipage}
\begin{minipage}[b]{0.32\linewidth}
\centering
\includegraphics[width=0.99\textwidth]{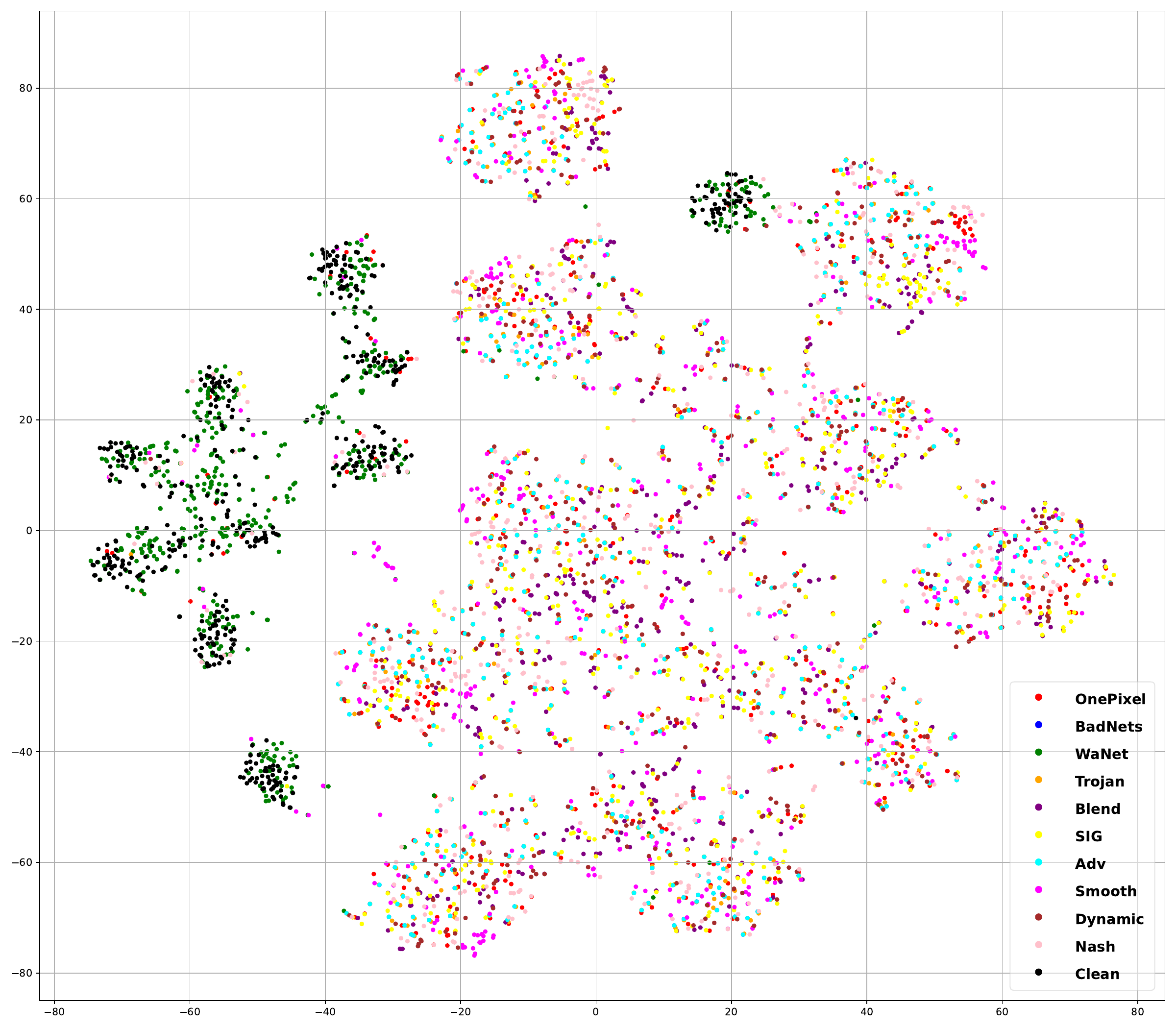}
\centerline{(b) All2All}
\end{minipage}
\begin{minipage}[b]{0.32\linewidth}
\centering
\includegraphics[width=0.99\textwidth]{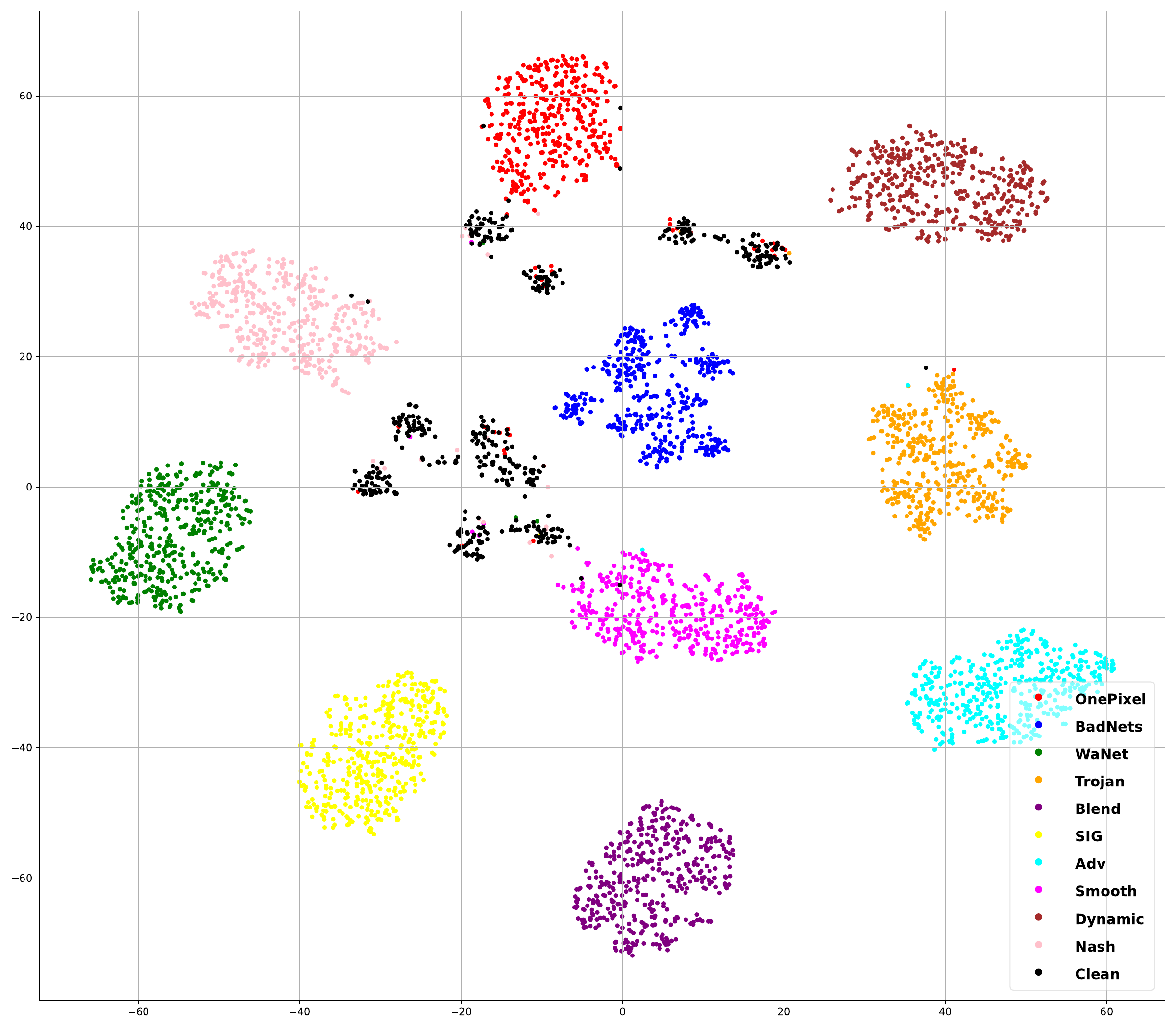}
\centerline{(c) All2Random}
\end{minipage}
\caption{The t-SNE visualizations of parallel MTBAs on ResNet-18 models trained on CIFAR-10. Each color represents the deep representations learned for one type of poisoned samples.}\label{fig:t-sne}
\vskip -0.20in
\end{figure}

\begin{figure}[!tp]
\small
\centering
\centerline{\includegraphics[width = .9\linewidth]{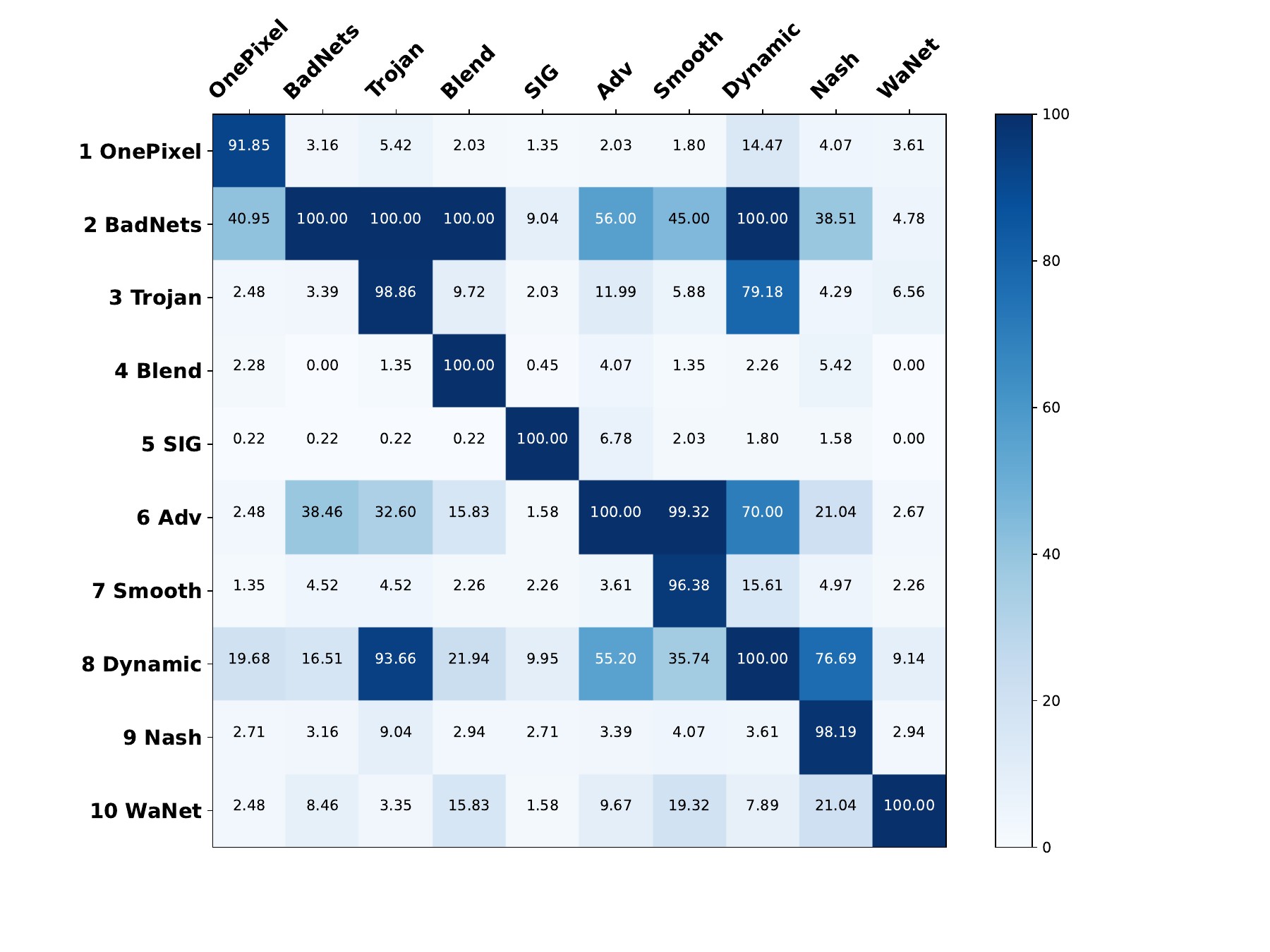}}
\vskip -0.10in
\caption{The ASR (\%) confusion matrix of sequential multi-trigger attacks on CIFAR-10 (ResNet-18). \emph{Row}: training triggers used to poison the model following the $1-9$ order; \emph{Column}: testing triggers used to activate the attack at test time.
}
\vskip -0.25in
\label{fig:confusion_matrix}
\end{figure}

\begin{figure*} 
\centering
\subfloat[Hybrid-trigger vs. its components] {\label{fig:hybrid-a}
\includegraphics[width=0.30\textwidth]{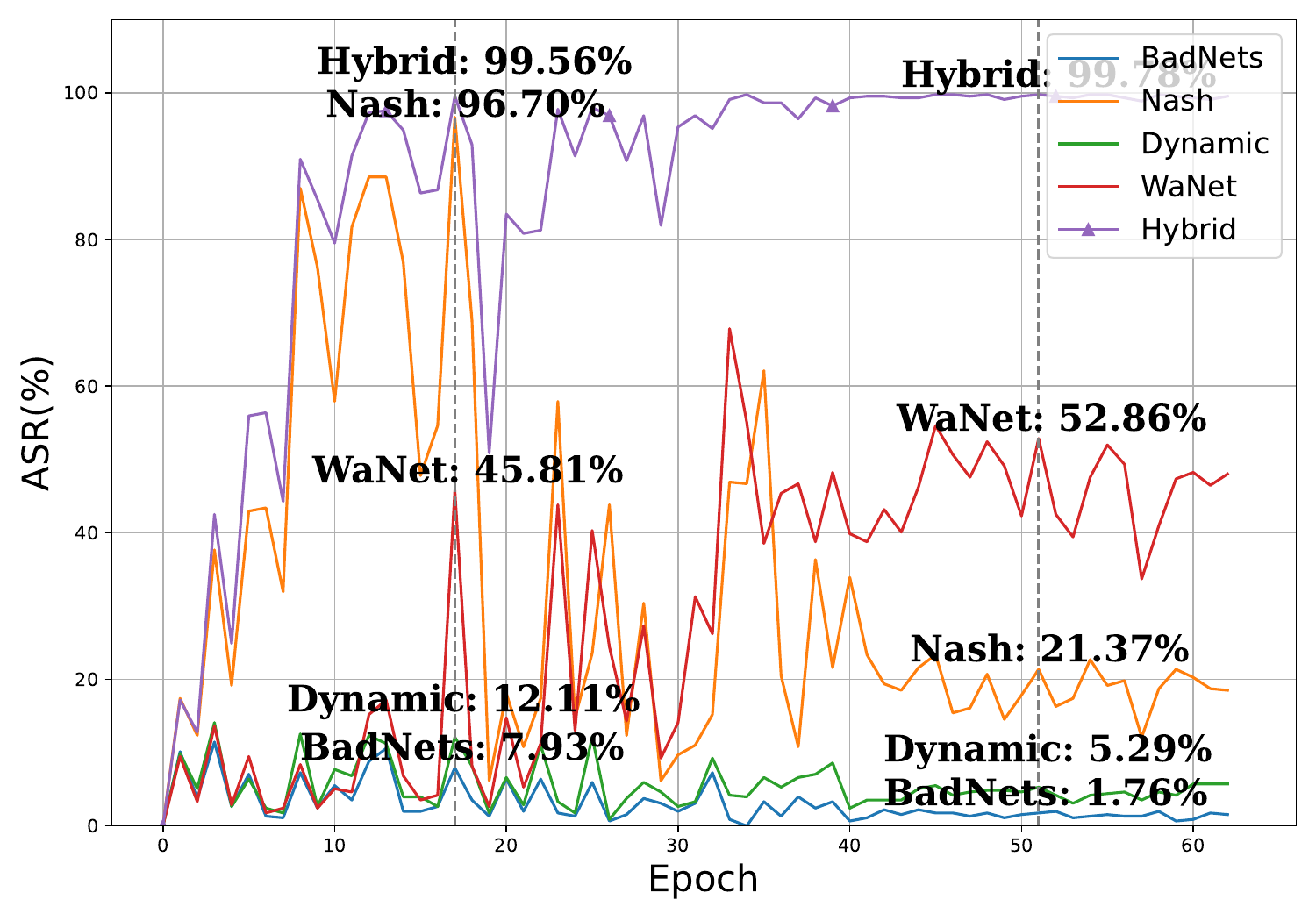}
}
\subfloat[Hybrid-trigger vs. irrelevant triggers] { \label{fig:hybrid-b}
\includegraphics[width=0.30\textwidth]{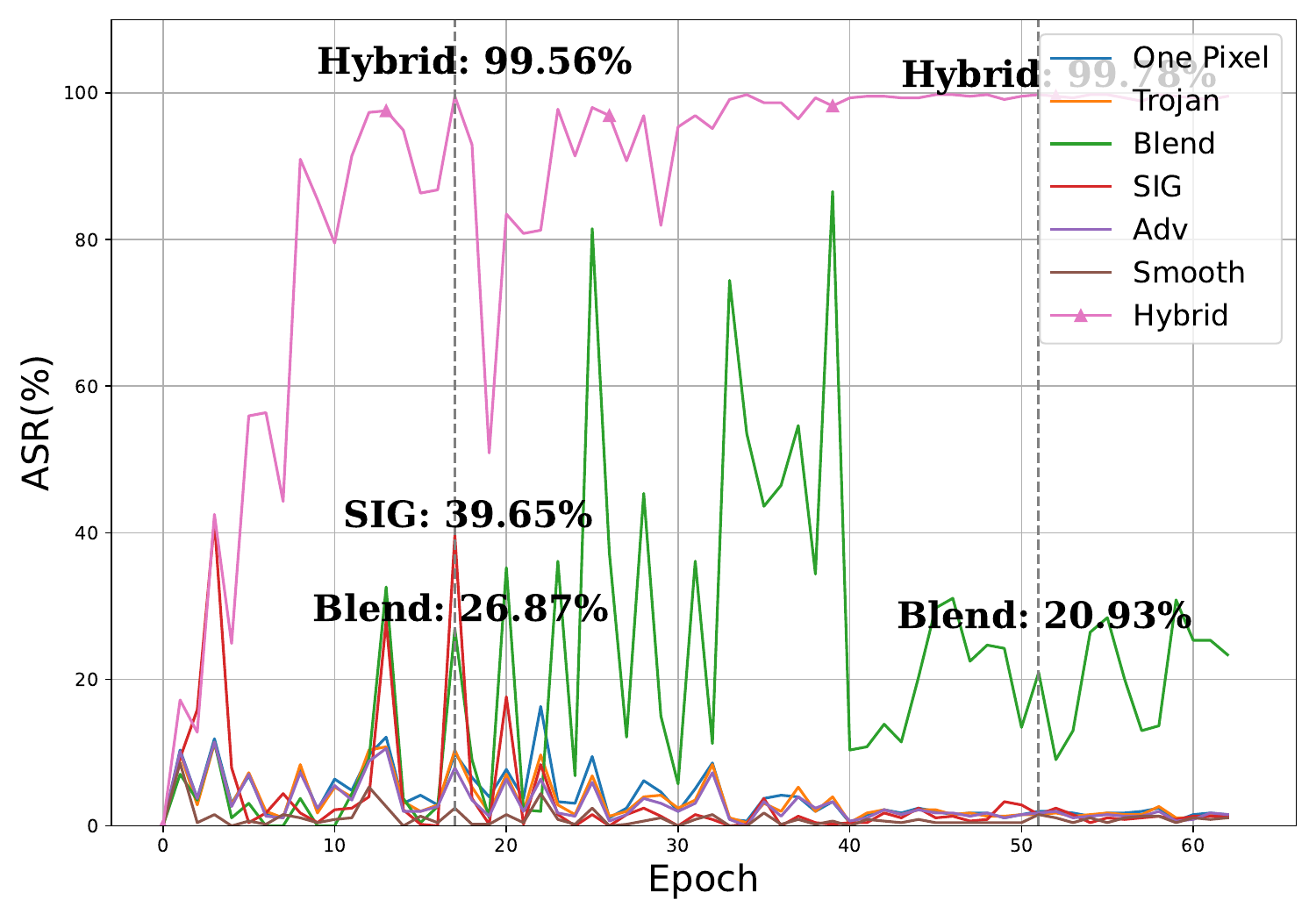}
}
\subfloat[Backdoor model detection] { \label{fig:detection-c}
\includegraphics[width=0.32\textwidth]{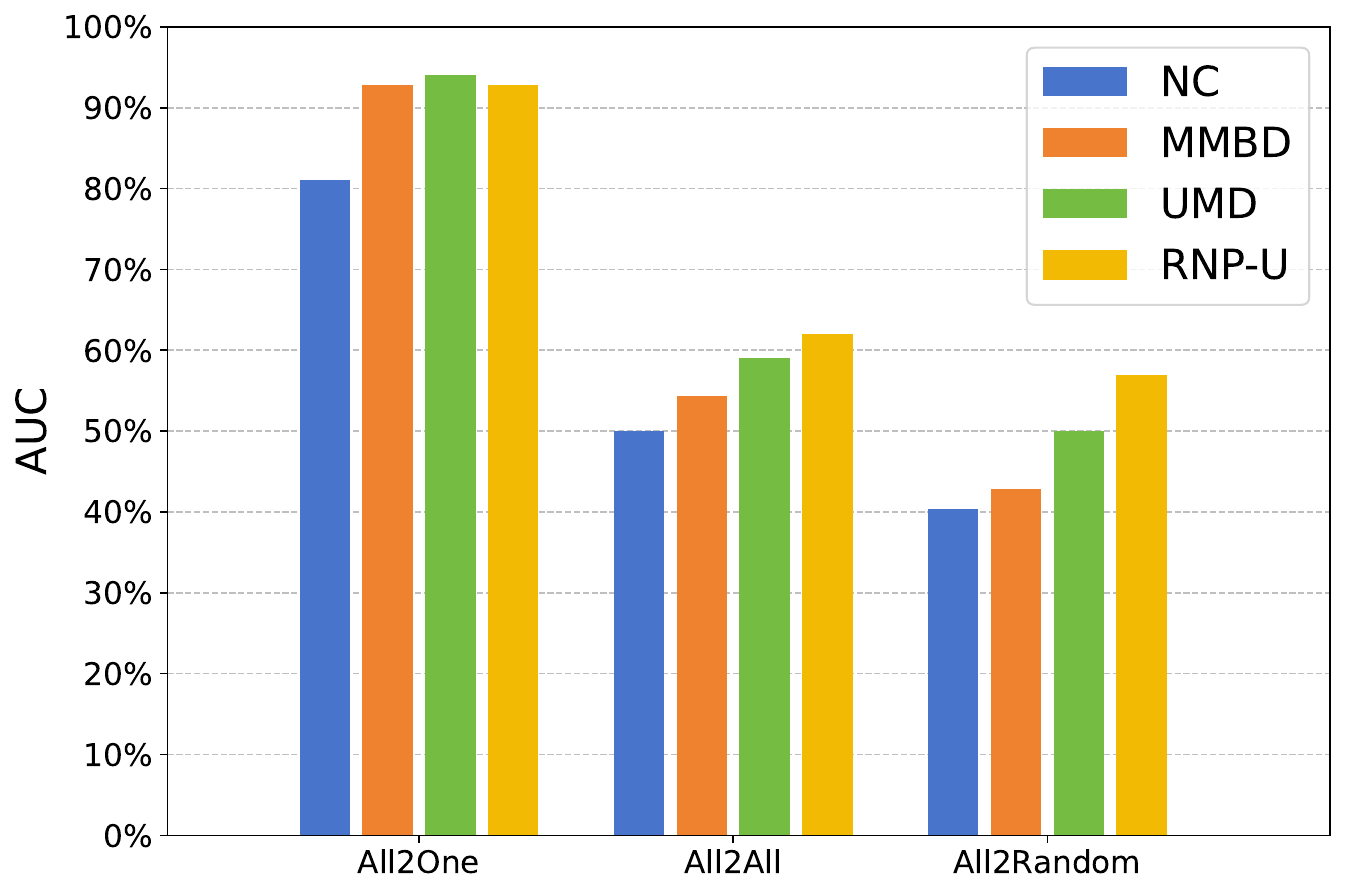}
}
\caption{The ASR (\%) of hybrid-trigger attack vs. its component triggers (a) and vs. other unrelated triggers (b), at different training epochs of the target model. (c) The backdoor model detection performance (AUROC) of different detection methods.}
\label{fig:hybrid}
\vskip -0.2in
\end{figure*}

Table \ref{tab:parallel_imagenet} displays the performance of parallel MTBAs on the ImageNet-20 subset. The experimental results are consistent with that on CIFAR-10. Particularly, most of the triggers can coexist with each other at a high poisoning rate (10\%), together achieving a high ASR above 80\%. Notably, with an increase in image scale and resolution, the ASR of full-image triggers (Blend, SIG, and Nash) surpasses that of the local triggers (BadNets and Trojan) at a low poisoning rate (1\%). We suspect that the full image triggers are somewhat enhanced on high-resolution images, rendering the backdoor features more prominent and easier to learn.

To help understand the clusters formed by different triggers, we show the t-SNE \cite{van2008visualizing} plots of the 10 triggers under the All2One, All2All, and All2Random label modifications in Fig. \ref{fig:t-sne}. It shows distinctive clustering effects under the three label modification strategies. Specifically, there exist both small compact clusters and relatively large clusters for All2One and All2Random attacks, yet no meaningful clusters for All2All attacks. This confirms that All2All attacks are indeed harder to achieve, highlighting the most complex attacking scenario. The independent clusters learned by the model for different triggers explain the coexisting effect between different triggers. \\

\subsubsection{Sequential MTBAs} \label{sec:seqential_mtba}

Sequential MTBAs involve the injection of the 10 triggers following a specified order for iterative training epochs (every 10 epochs). The injection order and the attack performance after each trigger was implanted into the model are illustrated in the form of a confusion matrix in Fig. \ref{fig:confusion_matrix}. As can be observed, there exists an \emph{overwriting effect} in sequential MTBAs, i.e., the effective trigger before became ineffective after new triggers were injected into the model. For instance, OnePixel drops from 91.85\% ASR to 40.95\% after BadNets was injected into the model, and further drops to 2.48\% after Trojan was injected. The same effect is also observed for almost all triggers, indicated by the consistent trend where the cells below the diagonal cells all have lower ASRs. \\

\noindent\textbf{Cross-activating Effect.} 
Another important observation is that certain rows have consistently higher ASRs (more blue cells), for example, the BadNets, Adv, and Dynamic rows. This is a \emph{cross-activating phenomenon}, that is, one type of trigger was frequently activated by other types of triggers. This supervising effect implies that the triggers share certain or even high similarities. For example, BadNets (the 2nd row in Fig. \ref{fig:confusion_matrix}) can be activated by BadNets, Torjan, Blend, Adv, Smooth, Dynamic, and Nash, with an ASR of 100\%, 100\%, 100\%, 56\%, 45\%, 100\%, and 38.51\%, respectively. The Dnaymic trigger can be activated by Trojan, Adv, Smooth, Dynamic, and Nash with an ASR of 93.66\%, 55.20\%, 35.74\%, 100\%, and 76.69\%, respectively. 
We speculate that the cross-activating effect is a result of similar pixel distributions in the trigger patterns, like those shared among BadNets, Trojan, and Dynamic triggers. 
Amongst the triggers, Trojan and Dynamic are the two most similar triggers, i.e., Trojan can activate Dynamic by an ASR of 93.66\% while Dynamic can activate Trojan by 79.18\%. This is because both triggers are optimized triggers. Additionally, BadNets' attack was reactivated by Dynamic attacks, increasing its ASR from 45\% back to 100\%. Similar observations apply to implicit triggers, such as Adv, Smooth, and Nash. The reasons behind these results may lie in the similarity of trigger patterns among different triggers. For instance, BadNets' static trigger pattern and Trojan attacks, as well as Dynamic attacks, shared the same pixel distribution, unintentionally preserving or reactivating the attack. We believe these subtle but intriguing relationships between different triggers deserve more in-depth investigation, especially under the multi-trigger setting. 

Although the threat level of sequential MTBAs is somewhat lower than that of parallel MTBAs, sequential attacks may be more suitable for distributed systems, such as federated learning, providing attackers with the conditions to implement such sequence attacks.

\begin{table*}[!tp]
\renewcommand{\arraystretch}{1.35} 
\centering
\caption{The performance (remaining ASR, \%) of 5 backdoor removal methods against the parallel multi-trigger attacks with total 10\% (trigger-wise 1\%) poisoning rate under the All2One, All2All, and All2Random modes. The experiments were done on CIFAR-10 using only 1\% (500) clean samples as the defense data.}
\begin{adjustbox}{width=0.9\linewidth}
\begin{tabular}{c|ccc|ccc|ccc|ccc|ccc|ccc}
\toprule
  & \multicolumn{3}{c|}{\textbf{No Defense}} 
  & \multicolumn{3}{c|}{\textbf{FT}} 
  & \multicolumn{3}{c|}{\textbf{FP}} 
  & \multicolumn{3}{c|}{\textbf{NAD}} 
  & \multicolumn{3}{c|}{\textbf{ANP}}
  & \multicolumn{3}{c}{\textbf{ABM}} \\ \cline{2-19} 
\multirow{-2}{*}{\textbf{Backdoor Removal}} 
  & All2One & All2All & All2Random 
  & All2One & All2All & All2Random 
  & All2One & All2All & All2Random 
  & All2One & All2All & All2Random 
  & All2One & All2All & All2Random
  & All2One & All2All & All2Random \\ \hline
\textbf{Clean Acc.} 
  & \textbf{93.98} & \textbf{93.44} & \textbf{93.2} 
  & \textbf{91.40} & \textbf{91.80} & \textbf{91.10} 
  & \textbf{83.00} & \textbf{87.60} & \textbf{86.80} 
  & \textbf{85.50} & \textbf{85.40} & \textbf{87.18} 
  & \textbf{85.30} & \textbf{86.60} & \textbf{86.40} 
  & \textbf{89.51} & \textbf{89.20} & \textbf{89.40} \\ \hline
OnePixel & 94.83 & 77.80 & 91.95 & 88.51 & 77.00 & 91.95 & 49.01 & 67.10 & 65.11 & 30.10 & 42.10 & 63.11 & 0.00 & 73.10 & 88.51 & 20.11 & 58.22 & 60.13 \\
BadNets  & 100.00 & 81.00 & 55.68 & 28.12 & 75.33 & 54.55 & 22.91 & 52.38 & 39.42 & 12.91 & 20.38 & 34.32 & 1.10 & 74.18 & 65.53 & 15.21 & 46.52 & 41.30 \\
Trojan   & 100.00 & 82.60 & 100.00 & 92.9 & 78.78 & 95.51 & 50.11 & 72.10 & 92.78 & 11.31 & 59.12 & 62.78 & 2.20 & 77.11 & 95.51 & 18.70 & 55.42 & 87.50 \\
Blend    & 99.56 & 77.20 & 100.00 & 99.12 & 25.13 & 100.00 & 78.81 & 74.30 & 96.13 & 32.09 & 38.32 & 56.13 & 11.00 & 22.33 & 72.53 & 24.33 & 40.10 & 64.70 \\
SIG      & 99.34 & 88.00 & 100.00 & 98.9 & 51.31 & 100.00 & 87.80 & 83.31 & 97.51 & 47.60 & 32.31 & 37.51 & 42.86 & 43.12 & 90.12 & 27.12 & 48.20 & 66.31 \\
Adv      & 100.00 & 81.00 & 39.41 & 19.22 & 75.52 & 31.82 & 43.40 & 42.22 & 43.52 & 23.20 & 68.32 & 43.52 & 13.10 & 74.62 & 83.55 & 29.80 & 44.20 & 58.12 \\
Smooth   & 93.40 & 81.20 & 94.62 & 83.52 & 72.78 & 95.70 & 69.23 & 57.80 & 86.28 & 22.75 & 65.80 & 86.28 & 1.10 & 91.00 & 98.92 & 31.25 & 50.14 & 71.20 \\
Nash     & 100.00 & 88.40 & 96.74 & 83.8 & 75.50 & 94.57 & 85.71 & 51.20 & 91.54 & 33.71 & 84.22 & 91.54 & 13.19 & 89.51 & 93.48 & 34.41 & 52.50 & 70.91 \\
Dynamic  & 97.37 & 89.66 & 100.00 & 97.6 & 81.11 & 100.00 & 97.10 & 78.93 & 99.11 & 68.41 & 67.38 & 99.11 & 7.69 & 65.53 & 100.00 & 28.91 & 49.61 & 79.81 \\
WaNet    & 97.59 & 77.60 & 100.00 & 95.3 & 69.13 & 98.72 & 66.79 & 63.21 & 89.33 & 25.78 & 57.48 & 54.53 & 5.14 & 64.92 & 3.04   & 26.71 & 42.32 & 62.30 \\  \hline
\rowcolor[HTML]{C0C0C0} 
\textbf{Average} 
  & \textbf{98.21} & \textbf{82.45} & \textbf{88.33} 
  & \textbf{79.85} & \textbf{70.31} & \textbf{86.28} 
  & \textbf{65.09} & \textbf{64.26} & \textbf{80.07} 
  & \textbf{30.79} & \textbf{53.54} & \textbf{62.88} 
  & \textbf{9.74}  & \textbf{67.54} & \textbf{79.12}
  & \textbf{23.67} & \textbf{49.30} & \textbf{68.05} \\ 
\bottomrule
\end{tabular}
\end{adjustbox}
\label{tab:removal}
\end{table*}

\vskip -0.15in
\subsection{Hybrid-trigger MTBAs} \label{sec:hybrid_mtba}
Different from parallel and sequential MTBAs where the triggers are independent, our hybrid-trigger attack mixes multiple triggers into a single clean sample, all pointing to the same backdoor target label.
This section explores the effectiveness of our hybrid-trigger attack on the CIFAR-10 dataset. To ensure high attack performance, we chose four triggers to construct the hybrid trigger, including one static trigger BadNets, one Dynamic, and two invisible triggers Nash and WaNet. Note that the Dynamic trigger does not work well as part of a hybrid. We apply those triggers in a random order (we will show that the order has minimum impact on the ASR) using a soft blending approach. 

Fig. \ref{fig:hybrid-a} illustrates the ASR of our hybrid-trigger attack and individual triggers. 
One key observation is that the hybrid trigger can not only achieve a high ASR ($\geq 90\%$) but also demonstrate a strong cross-activating effect on its component triggers (not all of them though). 
Particularly, besides the hybrid trigger, the ASR of Nash and WaNet reaches rapidly 96.70\% and 45.81\% respectively in the early stage of training (before 20 epochs), although they stabilize at 52.86\% and 21.37\% respectively in the end.  Figure \ref{fig:hybrid-b} shows a rather surprising phenomenon that the hybrid trigger can even cross-activate the Blend and SIG triggers in the middle of the training, even though they are not part of the hybrid trigger. Unfortunately, this cross-activation is temporary and disappears at the end of the training. We believe designing more advanced hybrid triggers that can cross-activate many other individual triggers is an interesting future direction.

\begin{figure}[!tp]
\centering
\begin{minipage}[t]{0.48\linewidth}
    \centering
    \includegraphics[width=\textwidth]{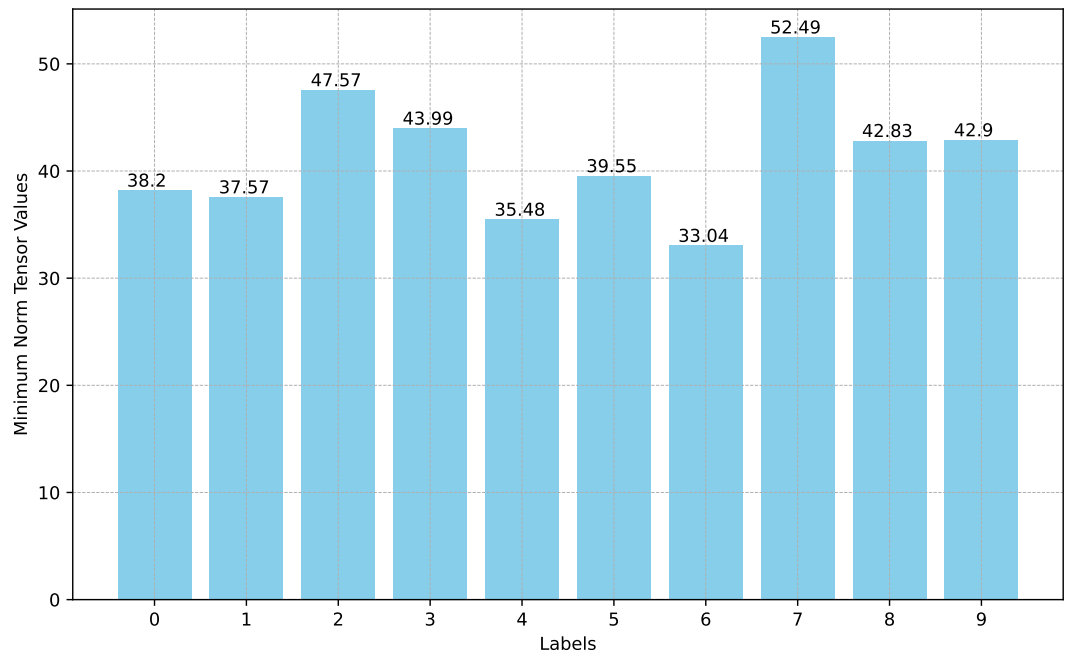}
    \caption*{(a) All2All}
\end{minipage}%
\hfill
\begin{minipage}[t]{0.48\linewidth}
    \centering
    \includegraphics[width=\textwidth]{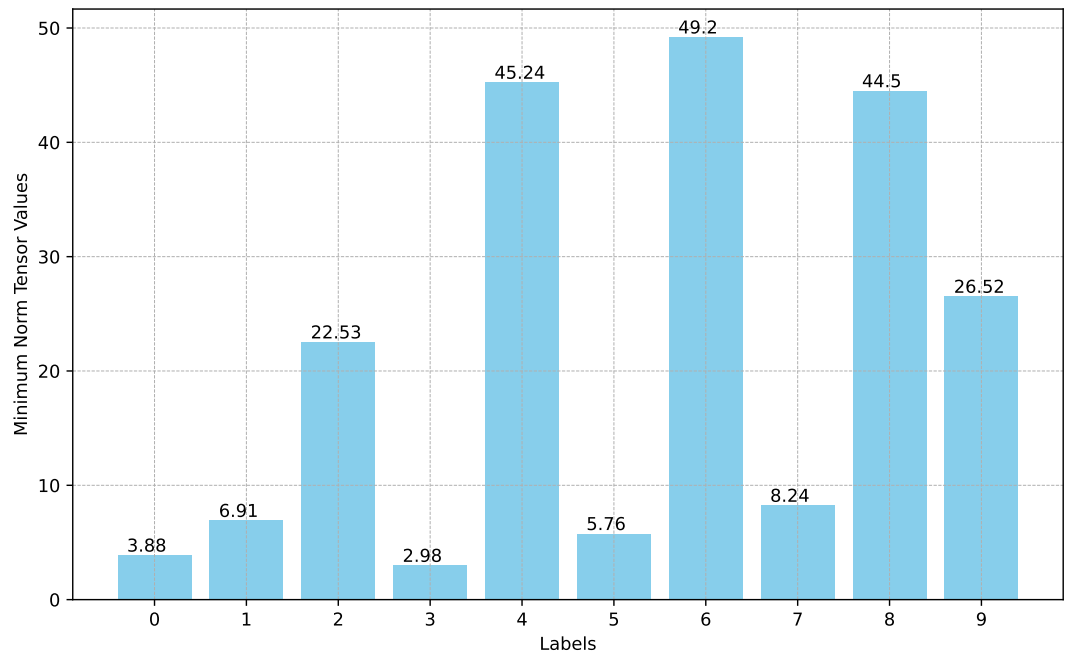}
    \caption*{(b) All2Random}
\end{minipage}
\caption{The L1-norm perturbation value for backdoor label detection under All2All and All2Rand settings in parallel MTBAs. The backdoored ResNet-18 model is trained on CIFAR-10 with a 10\% poisoning rate. We can find that NC struggles to infer the true backdoor label category as the perturbation values for all classes are inconsistent and random.}
\label{fig:NC}
\vspace{-0.2in}
\end{figure}

\vskip -0.15in
\subsection{Re-evaluating Existing Defenses}

This section examines the reliability of existing backdoor defense methods against multi-trigger attacks. Here, we re-evaluate 4 backdoor model detection methods and 4 backdoor removal methods. We only focus on parallel poisoning, i.e., the 10 triggers considered in this work were injected in parallel into the dataset. \\

\noindent\textbf{Backdoor Model Detection.}
Fig. \ref{fig:detection-c} presents the detection AUROC of 4 advanced backdoor model detection methods, including NC, MMBD, UMD, and RNP-U.
As can be observed, the four detection methods perform well on All2One multi-trigger attacks, yet perform poorly on All2All and All2Random attacks. Specifically, the detection AUROC deteriorates by nearly 50\% and 40\% on All2All and All2Random attacks respectively, when compared to that on All2One attacks. Such a huge degradation challenges the reliability of the four detection models under more complex scenarios with unfixed adversarial targets. Out of the four methods, RNP-U consistently outperforms the other three methods, which validates the usefulness of unlearning the benign features before detecting the backdoor features. Meanwhile, although these detection methods can detect the backdoor models to some extent, they all fail to identify the backdoor labels of All2All and All2Random attacks.

\noindent\textbf{Why do MTBAs bypass the existing backdoor detection methods?}
The primary issue with existing backdoor detection methods, such as NC, is their inability to effectively infer backdoor labels in our all-to-all (all2all) and all-to-random (all2rand) settings. In single-trigger scenarios, a significant statistical difference (e.g., absolute mean deviation) exists between backdoor labels and normal labels, allowing methods like NC to identify backdoor labels based on this statistical deviation. However, in MTBA settings, accurately inferring the true backdoor label becomes challenging when all labels are backdoored.

To validate this, we calculated the minimal perturbation value, i.e., L1-norm, required for most NC-like detection methods under both all2all and all2rand settings in parallel MTBAs. The backdoored ResNet-18 model was trained on CIFAR-10 with a 10\% poisoning rate. As illustrated in Fig. \ref{fig:NC}, the perturbation values across all classes exhibit inconsistency and randomness, making it difficult for NC to accurately infer the true backdoor label category.

\begin{table*}[!tp]
\renewcommand{\arraystretch}{1.25} 
\centering
\caption{The ASR/CA (\%) comparison between single-trigger attacks and parallel MTBAs. The results were obtained on CIFAR-10 with a poisoning rate of 10\%.}
\begin{adjustbox}{width=0.95\linewidth}
\begin{tabular}{c|c|cccccccccc|c}
\toprule
\textbf{Backdoor Attacks} & \textbf{Metrics} & \textbf{OnePixel} & \textbf{BadNets} & \textbf{Trojan} & \textbf{Blend} & \textbf{SIG} & \textbf{Adv} & \textbf{Smooth} & \textbf{Nash} & \textbf{Dynamic} & \textbf{WaNet} & \textbf{Average} \\ \hline
\multirow{2}{*}{Single-trigger} & CA & 92.69 & 93.7 & 92.99 & 93.7 & 93 & 93.19 & 91.69 & 93.66 & 92.99 & 93.51 & \textbf{93.11} \\
 & ASR & 96.62 & 100 & 98.18 & 100 & 92.34 & 99.75 & 95.56 & 98.12 & 99.97 & 99.1 & \textbf{97.96} \\ \hline
\multirow{2}{*}{Parallel MTBAs} & CA & 92.88 & 92.88 & 92.88 & 92.88 & 92.88 & 92.88 & 92.88 & 92.88 & 92.88 & 92.88 & \textbf{92.88} \\
 & ASR & 94.53 & 100 & 99.34 & 100 & 99.56 & 99 & 93.22 & 96.28 & 100 & 97.59 & \textbf{97.95} \\ 
\bottomrule
\end{tabular}
\end{adjustbox}
\label{tab:single2multi}
\vskip -0.15in
\end{table*}

\noindent\textbf{Backdoor Removal.} 
Table \ref{tab:removal} reports the defense performance of four backdoor removal methods, including FT, FP, NAD, ANP and ABM. We measure the defense performance by the remaining ASR, the lower the better. None of the defense methods can fully remove the multi-trigger from the model. The average ASRs are still well above 50\%, except for NAD and ANP against All2One attacks. 
Compared to NAD and ANP, fine-tuning and fine-pruning methods FT and FP are essentially ineffective against multi-trigger attacks, with an ASR $> 60\%$. 
This is because the presence of multiple triggers prevents effective forgetting of the backdoor correlation during standard fine-tuning. 
Additionally, the coexistence of multiple triggers leads to inconsistent parameter activations, rendering the activation-based pruning method FP ineffective.
The current state-of-the-art defense method ANP performs poorly on All2All and All2Random attacks, limiting its practicability against more realistic attacks that have different target labels. Clean accuracy (CA) is another perspective looking into the effectiveness of the defense method. 

The four backdoor removal methods cause a significant reduction in the model's clean accuracy when facing multi-trigger attacks. Specifically, against All2One attacks, FP, NAD, and ANP reduce the model's CA by 10.98\%, 8.48\%, and 8.68\% respectively, significantly impairing the functionality of the model. ABM reduces ASR to about 58.7\% on All2All while preserving clean accuracy above 89\%. However, its effectiveness is still far from sufficient under complex multi-trigger settings. Based on these results, we conclude that existing backdoor detection and removal methods struggle to address the threat of multi-trigger attacks, and developing more advanced countermeasures is imperative. \\


\begin{table*}[!tp]
\renewcommand{\arraystretch}{1.25} 
\centering
\caption{The ASR (\%) of parallel MTBAs across 4 different model architectures, i.e. ResNet-18, MobileNet-V2, VIT-Small, and VIT-Base, with the total/trigger-wise poisoning rate of 10\% (1\%). The experiment results were averaged on CIFAR-10. The best results are \textbf{boldfaced}.}
\begin{adjustbox}{width=0.95\linewidth}
\begin{tabular}{c|c|ccccccccccc}
\toprule
  &  & \multicolumn{11}{c}{\textbf{Parallel MTBAs (Poisoning rate 10\% (1\%))}} \\ \cline{3-13} 
\multirow{-2}{*}{\textbf{\begin{tabular}[c]{@{}c@{}}Label\\ Modification\end{tabular}}} & \multirow{-2}{*}{\textbf{Model}} & \multicolumn{1}{c|}{Clean Acc.} & OnePixel & BadNets & Trojan & Blend & SIG & Adv & Smooth & Nash & Dynamic & WaNet \\ \hline
 & ResNet-18 & \multicolumn{1}{c|}{92.88} & 94.53 & 100.00 & 99.34 & 100.00 & 99.56 & 99.00 & 93.22 & 96.28 & 100.00 & 97.59 \\
 & MobileNet-V2 & \multicolumn{1}{c|}{93.08} & 95.13 & 100.00 & 99.78 & 99.12 & 99.12 & 100.00 & 93.58 & 98.45 & 100.00 & 99.55 \\
 & VIT-Small & \multicolumn{1}{c|}{98.32} & 94.75 & 100.00 & 99.78 & 99.78 & 99.78 & 100.00 & 98.25 & 98.69 & 100.00 & 97.12 \\
 & VIT-Base & \multicolumn{1}{c|}{98.46} & 91.76 & 100.00 & 100.00 & 99.55 & 99.78 & 100.00 & 98.66 & 98.89 & 100.00 & 97.77 \\ \cline{2-13} 
\multirow{-5}{*}{All2One} & \cellcolor[HTML]{C0C0C0}\textbf{Average} & \multicolumn{1}{c|}{\cellcolor[HTML]{C0C0C0}\textbf{95.69}} & \cellcolor[HTML]{C0C0C0}\textbf{94.04} & \cellcolor[HTML]{C0C0C0}\textbf{100.00} & \cellcolor[HTML]{C0C0C0}\textbf{99.73} & \cellcolor[HTML]{C0C0C0}\textbf{99.61} & \cellcolor[HTML]{C0C0C0}\textbf{99.56} & \cellcolor[HTML]{C0C0C0}\textbf{99.75} & \cellcolor[HTML]{C0C0C0}\textbf{95.93} & \cellcolor[HTML]{C0C0C0}\textbf{98.08} & \cellcolor[HTML]{C0C0C0}\textbf{100.00} & \cellcolor[HTML]{C0C0C0}\textbf{98.01} \\ \hline
 & ResNet-18 & \multicolumn{1}{c|}{90.66} & 76.40 & 86.00 & 85.40 & 70.80 & 76.60 & 85.00 & 71.00 & 75.80 & 82.20 & 75.60 \\
 & MobileNet-V2 & \multicolumn{1}{c|}{92.96} & 86.20 & 92.80 & 91.40 & 76.00 & 83.60 & 91.80 & 82.20 & 84.20 & 86.80 & 84.40 \\
 & VIT-Small & \multicolumn{1}{c|}{98.26} & 86.80 & 98.20 & 97.60 & 90.40 & 95.20 & 97.20 & 90.80 & 94.00 & 96.80 & 90.60 \\
 & VIT-Base & \multicolumn{1}{c|}{98.54} & 83.60 & 98.00 & 98.40 & 94.00 & 96.00 & 97.10 & 91.20 & 93.00 & 96.80 & 88.00 \\ \cline{2-13} 
\multirow{-5}{*}{All2All} & \cellcolor[HTML]{C0C0C0}\textbf{Average} & \multicolumn{1}{c|}{\cellcolor[HTML]{C0C0C0}\textbf{95.11}} & \cellcolor[HTML]{C0C0C0}\textbf{83.25} & \cellcolor[HTML]{C0C0C0}\textbf{93.75} & \cellcolor[HTML]{C0C0C0}\textbf{93.20} & \cellcolor[HTML]{C0C0C0}\textbf{82.80} & \cellcolor[HTML]{C0C0C0}\textbf{87.85} & \cellcolor[HTML]{C0C0C0}\textbf{92.75} & \cellcolor[HTML]{C0C0C0}\textbf{83.80} & \cellcolor[HTML]{C0C0C0}\textbf{86.75} & \cellcolor[HTML]{C0C0C0}\textbf{90.65} & \cellcolor[HTML]{C0C0C0}\textbf{84.65} \\ \hline
 & ResNet-18 & \multicolumn{1}{c|}{92.00} & 89.01 & 43.01 & 100.00 & 100.00 & 98.90 & 47.73 & 88.17 & 95.56 & 100.00 & 96.51 \\
 & MobileNet-V2 & \multicolumn{1}{c|}{92.60} & 89.01 & 53.76 & 98.90 & 100.00 & 98.90 & 42.05 & 88.17 & 95.56 & 100.00 & 98.84 \\
 & VIT-Small & \multicolumn{1}{c|}{98.60} & 87.91 & 54.84 & 100.00 & 98.90 & 100.00 & 39.77 & 91.40 & 97.78 & 100.00 & 96.51 \\
 & VIT-Base & \multicolumn{1}{c|}{98.00} & 89.01 & 54.84 & 100.00 & 100.00 & 100.00 & 37.50 & 92.47 & 96.67 & 98.84 & 95.35 \\ \cline{2-13} 
\multirow{-5}{*}{All2Random} & \cellcolor[HTML]{C0C0C0}\textbf{Average} & \multicolumn{1}{c|}{\cellcolor[HTML]{C0C0C0}\textbf{95.30}} & \cellcolor[HTML]{C0C0C0}\textbf{88.74} & \cellcolor[HTML]{C0C0C0}\textbf{51.61} & \cellcolor[HTML]{C0C0C0}\textbf{99.73} & \cellcolor[HTML]{C0C0C0}\textbf{99.73} & \cellcolor[HTML]{C0C0C0}\textbf{99.45} & \cellcolor[HTML]{C0C0C0}\textbf{41.76} & \cellcolor[HTML]{C0C0C0}\textbf{90.05} & \cellcolor[HTML]{C0C0C0}\textbf{96.39} & \cellcolor[HTML]{C0C0C0}\textbf{99.71} & \cellcolor[HTML]{C0C0C0}\textbf{96.80} \\ \bottomrule
\end{tabular}
\end{adjustbox}
\label{tab:parallel_arch_0.1}
\end{table*}

\begin{figure*}[!tp]
\centering
\subfloat[BadNets, Dynamic, Nash, and WaNet] {\label{fig:hybrid_seq_a}
\includegraphics[width=0.30\textwidth]{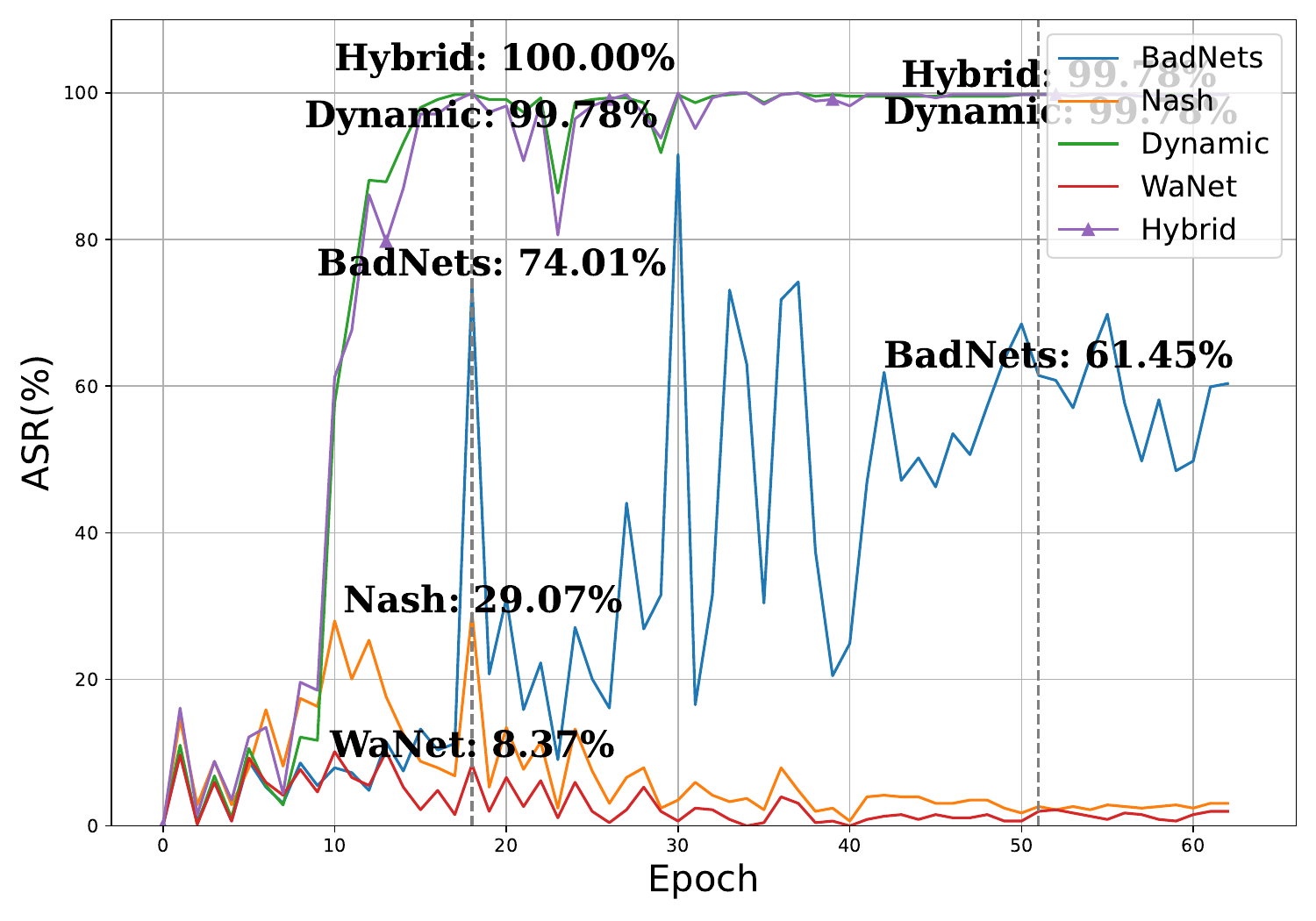}
}
\subfloat[BadNets, Nash, Dynamic, and WaNet] { \label{fig:hybrid_seq_b}
\includegraphics[width=0.30\textwidth]{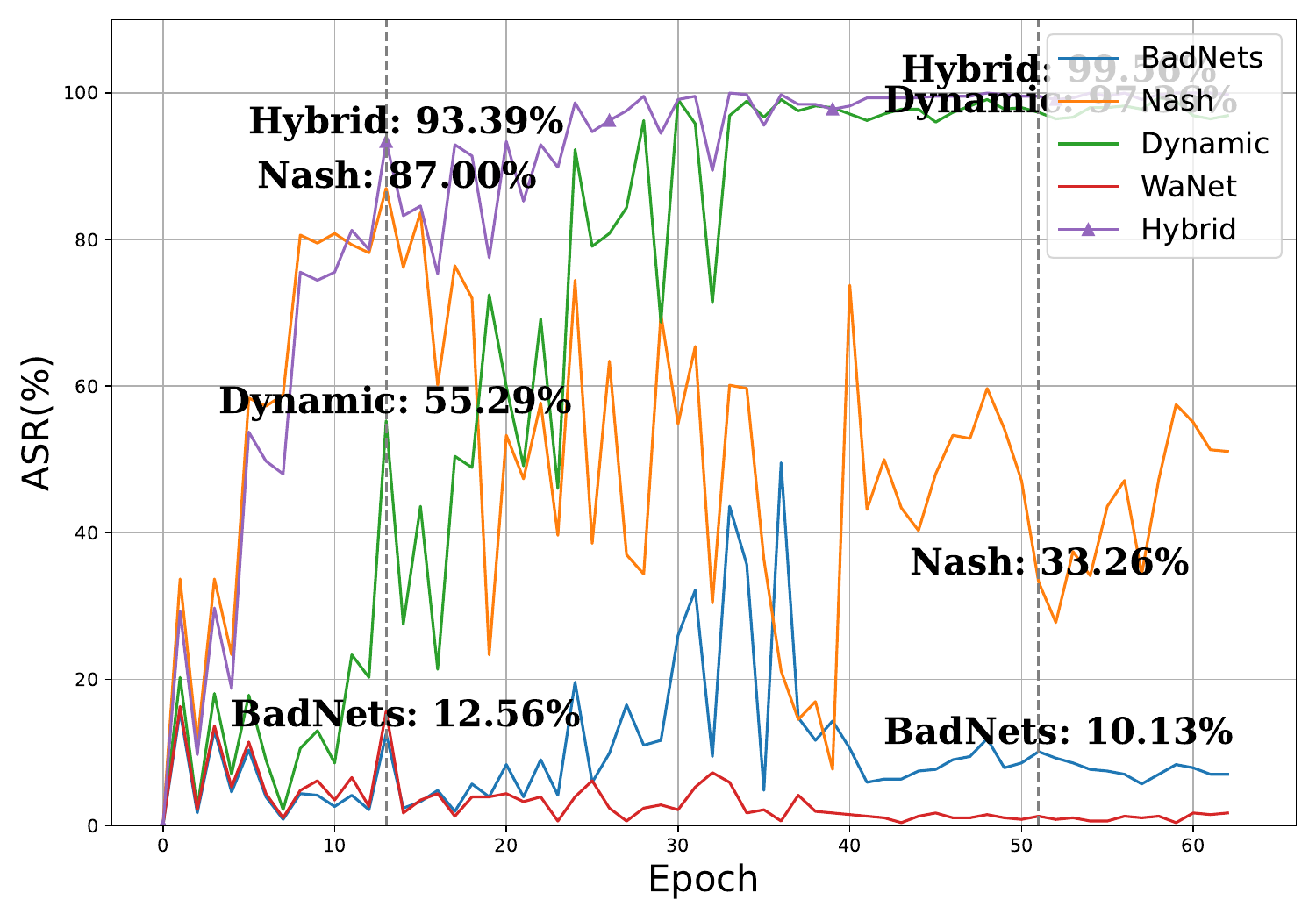}
}
\subfloat[Dynamic, WaNet, BadNets, and Nash] { \label{fig:hybrid_seq_c}
\includegraphics[width=0.30\textwidth]{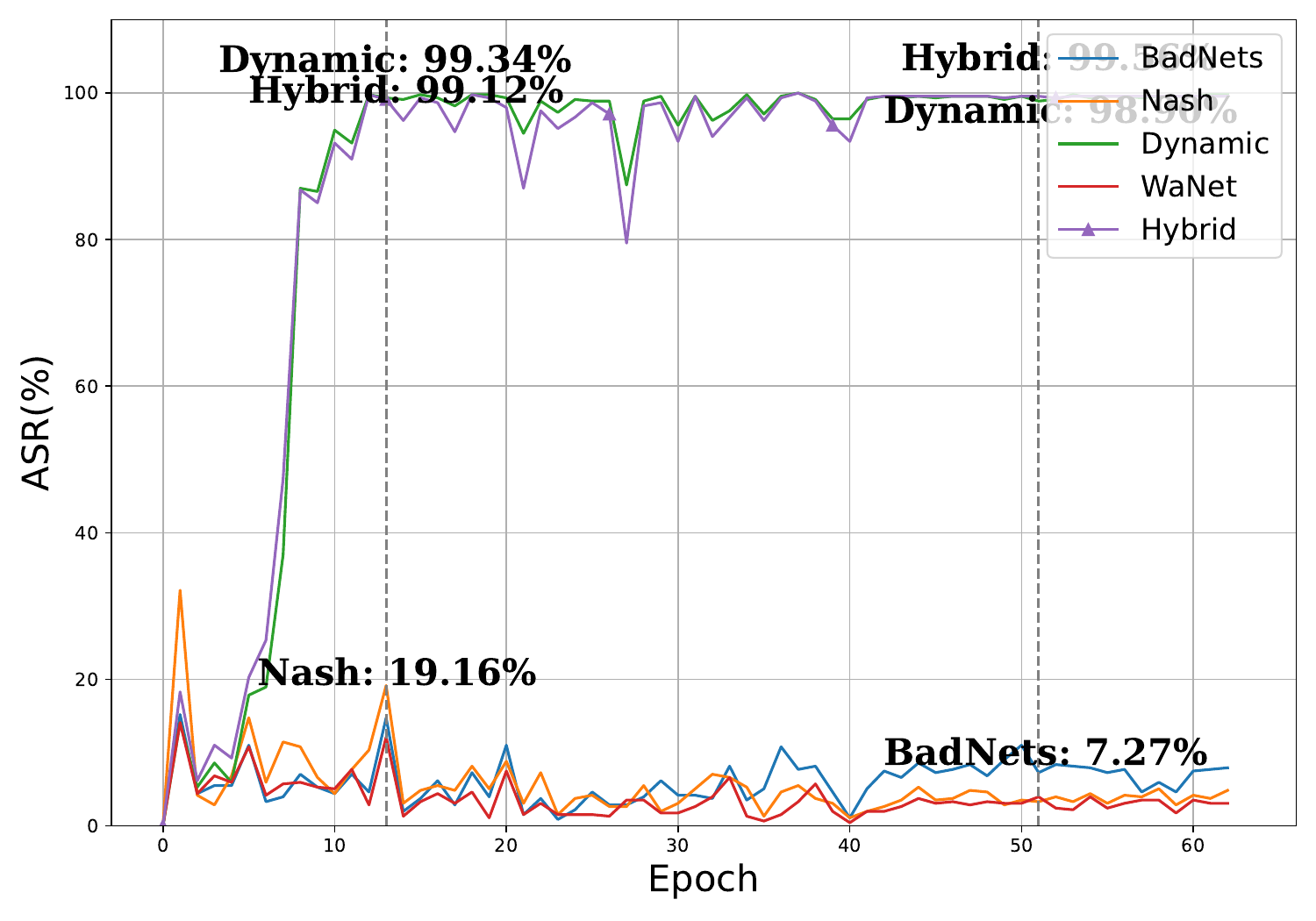}
}
\caption{Hybrid-trigger attacks with different trigger stacking orders.}
\label{fig:hybrid_seq}
\vskip -0.15in
\end{figure*}

\vskip -0.2in
\subsection{Analysis and Further Exploration}

\noindent\textbf{Single-Trigger vs. Multi-Trigger Attacks.}  
Table \ref{tab:single2multi} presents the comparative performance of single-trigger and multi-trigger (Parallel MTBA) attacks on a CIFAR-10 dataset with a ResNet-18 model and a consistent poisoning rate of 10\%. The results indicate that multi-trigger attacks achieve nearly identical Attack Success Rates (ASR) and Clean Accuracy (CA) compared to single-trigger attacks. For example, the average CA for single-trigger attacks is 93.11\%, while it is 92.88\% for multi-trigger attacks, showing only a 0.23\% difference. Similarly, the average ASR for single-trigger attacks is 97.96\%, compared to 97.95\% for multi-trigger attacks, reflecting a negligible 0.01\% difference. These values suggest that introducing multiple triggers does not significantly degrade the model’s performance on clean data or the effectiveness of the attack.

In terms of specific cases, single-trigger attacks with the BadNets method achieve an ASR of 100\% and a CA of 93.7\%, while multi-trigger attacks with BadNets also reach an ASR of 100\% with a slightly lower CA of 92.88\%. Another example is the Dynamic attack, where the single-trigger ASR is 99.97\% and the CA is 92.99\%, closely matched by the multi-trigger ASR of 100\% and CA of 92.88\%. These similarities across various methods reinforce that multi-trigger attacks can maintain high ASR and CA values, effectively balancing attack strength and model utility.

The minimal impact on CA suggests that models compromised with multiple triggers remain highly functional for benign tasks, making detection more challenging. The ability of multi-trigger attacks to maintain similar ASR to single-trigger attacks, despite involving multiple triggers, highlights their robustness and adaptability across diverse attack scenarios. This increased attack diversity allows multi-trigger methods to be more flexible and resilient, potentially posing a more significant threat in real-world applications where detection is crucial. \\

\noindent\textbf{Parallel MTBAs Across different Architectures.}  
Table \ref{tab:parallel_arch_0.1} reports the attack performance of parallel MTBAs on four distinct model architectures, including two Convolutional Neural Networks (CNNs) (i.e. ResNet-18 and MobileNet-V2) and two Vision Transformers (ViTs) (i.e. VIT-Small and VIT-Base), at a poisoning rate of 10\%. The results indicate that MTBAs achieve high ASR and maintain CA across different architectures, demonstrating a high degree of generalizability.

Under the All2One label modification, ResNet-18 achieves a clean accuracy of 92.88\% with ASR values of 100\% for both BadNets and Dynamic attacks. Similarly, VIT-Base achieves a clean accuracy of 98.46\% and reaches 100\% ASR with the BadNets, Trojan, and Dynamic methods. These high ASR values across both CNN and ViT architectures suggest that parallel MTBAs are highly effective across different model structures without significantly impacting clean accuracy, with an average CA of 95.69\% for All2One label modification. However, the effectiveness of certain attacks varies slightly based on architecture. For example, in the All2All label modification, ResNet-18 shows lower ASR values for attacks like OnePixel (76.40\%) and SIG (76.60\%), compared to MobileNet-V2, which achieves higher ASR values of 86.20\% for OnePixel and 83.60\% for SIG. This variation may be attributed to differences in architectural sensitivity to small perturbations, as CNNs and ViTs process input data differently, with ViTs relying on token-based attention mechanisms that may slightly reduce the influence of certain types of injected triggers.

These findings highlight the versatility of MTBAs in compromising various architectures while maintaining high clean accuracy, especially at higher poisoning rates. The consistent effectiveness of attacks across architectures underscores the need for adaptive defense mechanisms that can detect and mitigate MTBAs in both CNNs and ViTs, accommodating differences in architectural vulnerabilities and the persistence of high ASR values. \\

\noindent\textbf{Hybrid-Trigger MTBAs with Different Stacking Orders.}  
To further explore the impact of hybrid-trigger attacks, we analyzed how different stacking orders of four trigger patterns—BadNets, Dynamic, Nash, and WaNet—affect their cross-activation capabilities. As shown in Figures \ref{fig:hybrid_seq}(a)-(c), our analysis reveals two important observations: 1) the order of trigger stacking influences the overall cross-activating effect, where triggers applied earlier may be compromised or overwritten by subsequent ones, diminishing their effectiveness; and 2) the Dynamic trigger consistently demonstrates stronger cross-activating capabilities compared to the other patterns, maintaining a high ASR across various stacking sequences. This indicates that certain triggers, such as Dynamic, possess intrinsic properties that make them more resistant to interference from other triggers, thereby sustaining their attack potential. These findings emphasize the importance of trigger selection and ordering in hybrid-trigger attacks, as the interplay between triggers can significantly impact the overall effectiveness of the attack.

These hybrid triggers present a significant challenge for traditional backdoor detection methods, as the combination of multiple trigger patterns complicates both their identification and mitigation. This underscores the need for advanced defense mechanisms that can handle complex, multi-trigger backdoor attacks and adapt to their varying stacking strategies. In summary, the analysis highlights the sophistication and adaptability of multi-trigger backdoor attacks, as well as the critical need for robust, architecture-aware defenses to counteract them effectively.

\section{Discussion}
This work introduces three poisoning strategies for multi-trigger backdoor attacks (MTBAs): \textbf{parallel}, \textbf{sequential}, and \textbf{hybrid}, covering scenarios involving both independent and collusive adversaries. Beyond the empirical findings, several important insights emerge that can guide future research on understanding and defending against MTBAs.

\subsubsection{Potential Directions for Hybrid Defense Mechanisms}
Existing defenses are typically designed to operate at a single stage of the backdoor pipeline, such as backdoor unlearning or backdoor removal. However, combining techniques across stages may offer stronger protection. We conducted preliminary evaluations that integrate backdoor unlearning (BU) and removal-based methods (BR). Results show that hybrid approaches—such as combining BU with fine-tuning (BU + FT), adversarial neuron pruning (BU + ANP), or anomaly-based learning (BU + ABL)—achieve significantly lower attack success rates (ASR $\leq$ 3\%) while preserving clean accuracy above 90\%. This suggests that chaining complementary defenses, e.g., exposure via unlearning followed by backdoor removal, can be more effective than using any single method alone.

\subsubsection{Towards a Theoretical Understanding of Trigger Interactions}
While our work primarily emphasizes empirical evaluation, a deeper theoretical understanding of multi-trigger attacks remains both critical and underexplored in backdoor research. We highlight three preliminary perspectives that shed light on why and how triggers interact: (1) \emph{Representational dominance}, where salient triggers with stronger and more visible features (e.g., BadNets, Dynamic) dominate the optimization process, thereby overwriting weaker and stealthier triggers (e.g., Nash, WaNet); (2) \emph{Shared feature representation}, where multiple triggers overlap in representational subspaces, causing their gradient signals to interfere and jointly activate shared neurons, leading to cross-activation effects; and (3) \emph{Parametric memory}, where residual traces of previously learned triggers persist in the weight space even after being suppressed, allowing them to partially re-emerge when reinforced by correlated patterns.

Providing a rigorous theoretical analysis of these mechanisms is both important and challenging, and we acknowledge that this aspect remains underexplored in the backdoor literature. As part of our future work, we plan to investigate these dynamics in greater depth using information-theoretic and spectral tools (e.g., mutual information, singular value analysis), with the goal of establishing a more principled foundation for understanding multi-trigger backdoors.


\section{Conclusion}

This work promotes the concept of multi-trigger backdoor attacks (MTBAs) where multiple adversaries may use different types of triggers to poison the same dataset. By designing three poisoning strategies, including parallel, sequential, and hybrid-trigger poisonings, we conducted extensive experiments to study the properties of multi-trigger attacks with 10 representative triggers. Through these experiments, we revealed the coexisting, overwriting, and cross-activating effects of multi-trigger attacks. We also showed the limitations of existing defense methods in detecting and removing multiple triggers from a backdoored model. We hope our work can help reshape future backdoor research towards more realistic settings.


\bibliographystyle{IEEEtran}
\bibliography{IEEEabrv,reference}


\newpage

%


\begin{IEEEbiography}[{\includegraphics[width=1in,height=1.25in,clip,keepaspectratio]{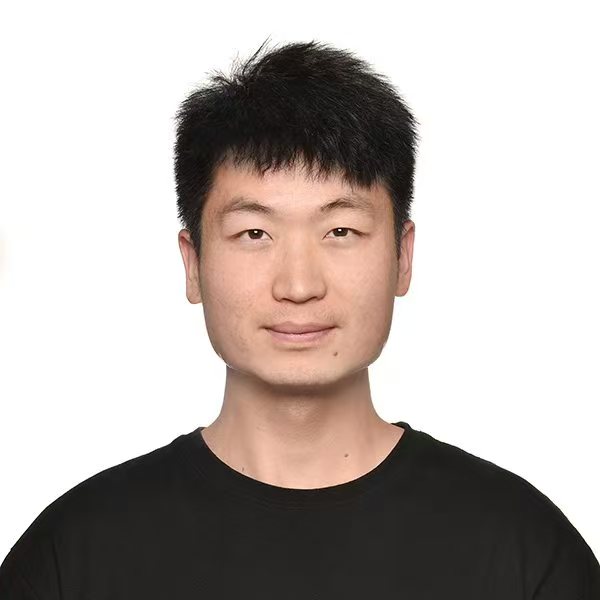}}]{Yige Li}
is a Postdoctoral Research Fellow at Singapore Management University. He received his Ph.D. from Xidian University. His research interests include generative models (LLMs, VLMs, and Agents) and AI safety, with a focus on developing simple yet insightful solutions grounded in theory. He serves as a PC member for top conferences such as ICLR, NeurIPS, and ICML, and as a reviewer for leading journals including IEEE TPAMI and IEEE TDSC.
\end{IEEEbiography}

\begin{IEEEbiography}[{\includegraphics[width=1in,height=1.25in,clip,keepaspectratio]{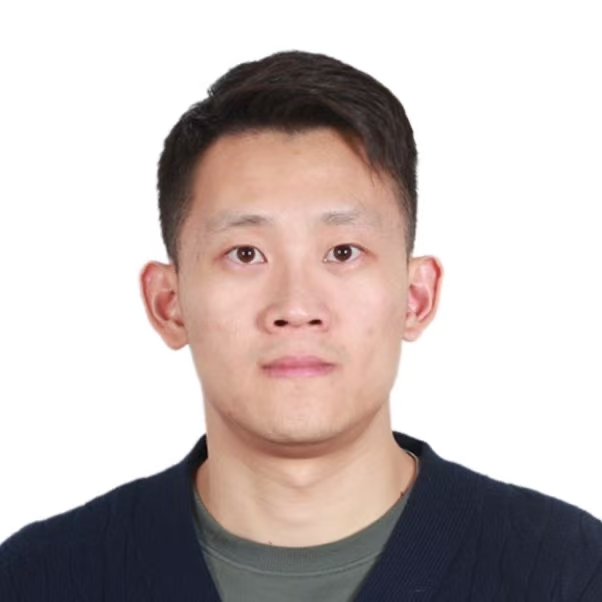}}]{Jiabo He}
is currently an AI scientist at Bosch Research Asia Pacific \& Bosch Center for Artificial Intelligence (BCAI). He received his Ph.D. degree from The University of Melbourne in 2022. His main research area is Computer Vision, Multi-modal Large Language Model (MLLM), and trustworthy AI for practical applications. He has published 10+ top-tier conference and journal papers, as well as books and patents.
\end{IEEEbiography}

\begin{IEEEbiography}[{\includegraphics[width=1in,height=1.25in,clip,keepaspectratio]{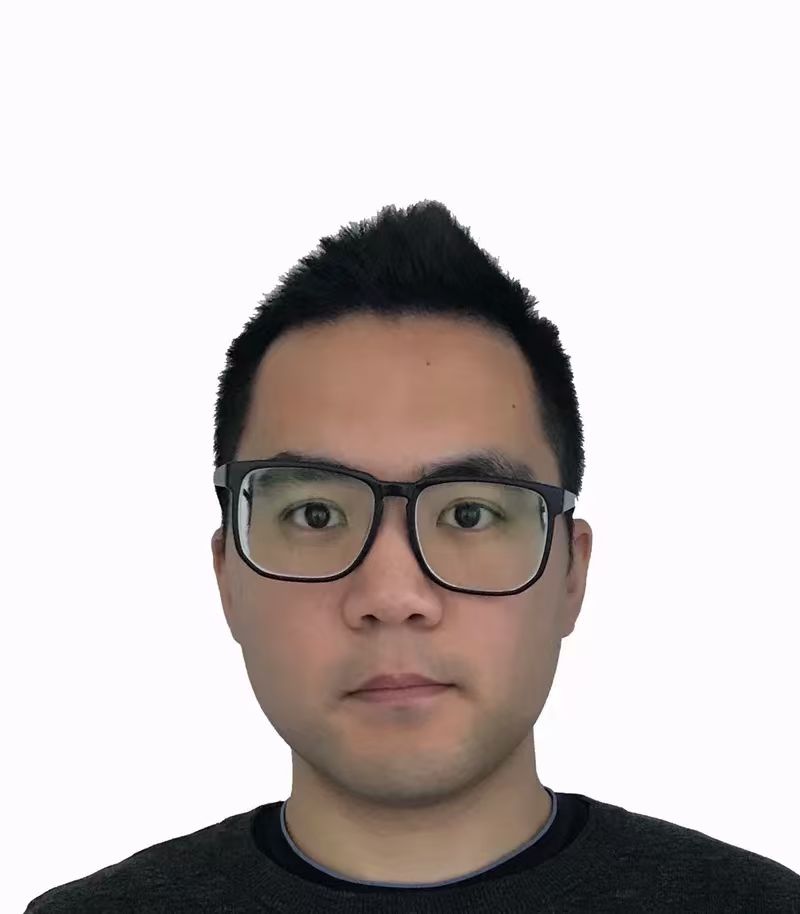}}]{Hanxun Huang}
is currently a Postdoctoral Research Fellow at the School of Computing and Information Systems, The University of Melbourne. He received his PhD degree from the University of Melbourne. His main research area is trustworthy AI, aiming to design secure, robust, privacy-preserving, and fair machine learning models for real-world AI applications. He has published at top-tier conferences and journals, including ICLR, NeurIPS, and ICML.
\end{IEEEbiography}

\begin{IEEEbiography}[{\includegraphics[width=1in,height=1.25in,clip,keepaspectratio]{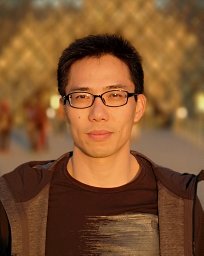}}]{Jun Sun}
is a Professor at Singapore Management University (SMU). He received his B.S. and Ph.D. degrees in Computer Science from the National University of Singapore (NUS) in 2002 and 2006, respectively, and has been a faculty member since 2010. His research interests include AI safety, software engineering, and formal methods. He enjoys designing algorithms as well as life beyond research. He has been awarded the Lee Kuan Yew Fellowship twice, and has published extensively in CCF-A conferences and journals, with multiple ACM Distinguished Paper Awards.
\end{IEEEbiography}

\begin{IEEEbiography}[{\includegraphics[width=1in,height=1.25in,clip,keepaspectratio]{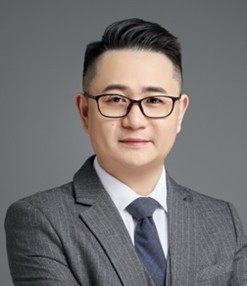}}]{Xingjun Ma}
is currently an associate professor at Fudan University. He received his PhD degree from The University of Melbourne. His main research area is trustworthy AI, aiming to design secure, robust, explainable, privacy-preserving, and fair machine learning models for real-world AI applications. He has published at top-tier conferences and journals, including ICLR, NeurIPS, and ICML. He also serves as an area chair for ICLR and ICML. 
\end{IEEEbiography}

\begin{IEEEbiography}[{\includegraphics[width=1in,height=1.25in,clip,keepaspectratio]{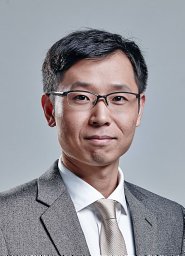}}]{Yu-Gang Jiang}
(Fellow, IEEE) received the Ph.D. degree in Computer Science from City University of Hong Kong in 2009 and worked as a Postdoctoral Research Scientist at Columbia University, New York, during 2009–2011. His research lies in the areas of multimedia, computer vision, embodied AI, and trustworthy AI. His research has led to the development of innovative AI tools that have been used in many practical applications like defect detection for high-speed railway infrastructures. His open-source video analysis toolkits and datasets such as CU-VIREO374, CCV, THUMOS, FCVID, and WildDeepfake have been widely used in both academia and industry. He currently serves as Chair of ACM Shanghai Chapter and Associate Editor of several international journals. For contributions to large-scale and trustworthy video analysis, he was elected Fellow of IEEE, IAPR, and CCF.
\end{IEEEbiography}




\end{document}